\documentclass[10pt,twocolumn,letterpaper]{article}

\usepackage[pagenumbers]{cvpr} 

\usepackage{graphicx}
\usepackage{amsmath}
\usepackage{amssymb}
\usepackage{booktabs}
\usepackage{tabularx}
\usepackage{multirow}
\newcolumntype{C}{>{\centering\arraybackslash}X} 
\usepackage{arydshln}
\usepackage{bbm}
\usepackage{enumitem}            
\usepackage{bm}
\usepackage[accsupp]{axessibility}

\usepackage[pagebackref,breaklinks,colorlinks]{hyperref}
\usepackage{xcolor}

\setlist[itemize]{noitemsep,topsep=0pt,leftmargin=*,label={\large\textbullet}}
\newcommand{\Section}[1]{\vspace{-1mm} \section{#1} \vspace{-1mm}}

\newcommand{\Paragraph}[1]{\vspace{1.1mm}\noindent\textbf{#1.}\hspace{0.5mm}}

\newcommand{\first}[1]{\textcolor{red}{\bm{#1}}}
\newcommand{\second}[1]{\textcolor{blue}{\bm{#1}}}
\newcommand{\firstkey}[1]{\textcolor{red}{\textbf{#1}}}
\newcommand{\secondkey}[1]{\textcolor{blue}{\textbf{#1}}}

\usepackage[capitalize]{cleveref}
\crefname{section}{Sec.}{Secs.}
\Crefname{section}{Section}{Sections}
\Crefname{table}{Table}{Tables}
\crefname{table}{Tab.}{Tabs.}

\def\cvprPaperID{6682} 
\def\confName{CVPR}

\begin{document}

\title{AdaFace: Quality Adaptive Margin for Face Recognition}

\author{Minchul Kim, Anil K. Jain,  Xiaoming Liu\\
Department of Computer Science and Engineering,\\
Michigan State University, East Lansing, MI, 48824\\
{\tt\small{\{kimminc2, jain, liuxm\}@cse.msu.edu}}}

\maketitle

\begin{abstract}
Recognition in low quality face datasets is challenging because facial attributes are obscured and degraded. Advances in margin-based loss functions have resulted in enhanced discriminability of faces in the embedding space. Further, previous studies have studied the effect of adaptive losses to assign more importance to misclassified (hard) examples. In this work, we introduce another aspect of adaptiveness in the loss function, namely the image quality. We argue that the strategy to emphasize misclassified samples should be adjusted according to their image quality. Specifically, the relative importance of easy or hard samples should be based on the sample's image quality. We propose a new loss function that emphasizes samples of different difficulties based on their image quality. Our method achieves this in the form of an adaptive margin function by approximating the image quality with feature norms. Extensive experiments show that our method, AdaFace, improves the face recognition performance over the state-of-the-art (SoTA) on four datasets (IJB-B, IJB-C, IJB-S and TinyFace). Code and models are released in \href{https://github.com/mk-minchul/AdaFace}{Supp}.

\end{abstract}

\section{Introduction}
\label{sec:intro}
\begin{figure}[t]
\centering
  \includegraphics[width=0.95\linewidth]{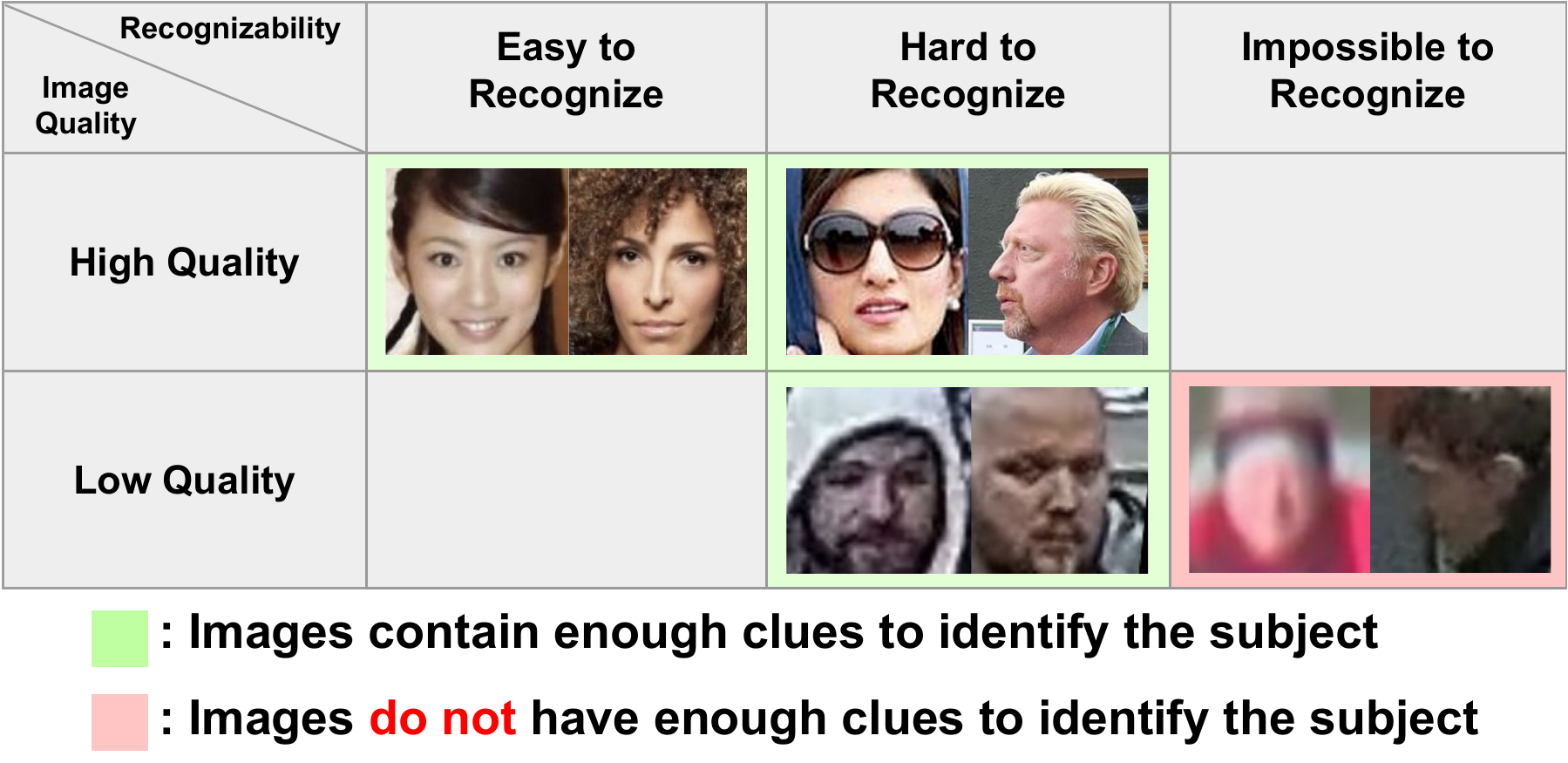}
  \vspace{-1mm}
  \caption{
  Examples of face images with different qualities and recognizabilities.
  Both high and low quality images contain variations in pose, occlusion and resolution that sometimes make the recognition task difficult, yet achievable. 
  Depending on the degree of degradation, some images may become impossible to recognize. By studying the different impacts these images have in training, this work aims to design a novel loss function that is adaptive to a sample's recognizability, driven by its image quality. \vspace{-4mm}}
  \label{figure:concept_figure}
\end{figure}

Image quality is a combination of attributes that indicates how faithfully an image captures the original scene~\cite{sheikh2006image}. Factors that affect the image quality include brightness, contrast, sharpness, noise, color constancy, resolution, tone reproduction, \textit{etc.}  Face images, the focus of this paper, can be captured under a variety of settings for lighting, pose and facial expression, and sometimes under extreme visual changes such as a subject's age or make-up. These parameter settings make the recognition task difficult for learned face recognition (FR) models. Still, the task is achievable in the sense that humans or models can often recognize faces under these difficult settings~\cite{disentangled-representation-learning-gan-for-pose-invariant-face-recognition}. However, when a face image is of low quality, depending on the degree, the recognition task becomes infeasible. Fig.~\ref{figure:concept_figure} shows examples of both high quality and low quality face images. It is not possible to recognize the subjects in the last column of Fig.~\ref{figure:concept_figure}. 

Low quality images like the bottom row of Fig.~\ref{figure:concept_figure} are increasingly becoming an important part of face recognition datasets because they are encountered in surveillance videos and drone footage. Given that SoTA FR methods~\cite{deng2019arcface, huang2020curricularface, li2021spherical, deng2021variational} are able to obtain over 98\% verification accuracy in relatively high quality datasets such as LFW or CFP-FP~\cite{lfw, cfpfp}, recent FR challenges have moved to lower quality datasets such as IJB-B, IJB-C and IJB-S~\cite{ijbb, ijbc, ijbs}. Although the challenge is to attain high accuracy on low quality datasets, most popular training datasets still remain comprised of high quality images~\cite{msceleb, deng2019arcface}. Since only a small portion of training data is low quality, it is important to properly leverage it during training.

\begin{figure*}[t]
    \centering
    \includegraphics[width=\linewidth]{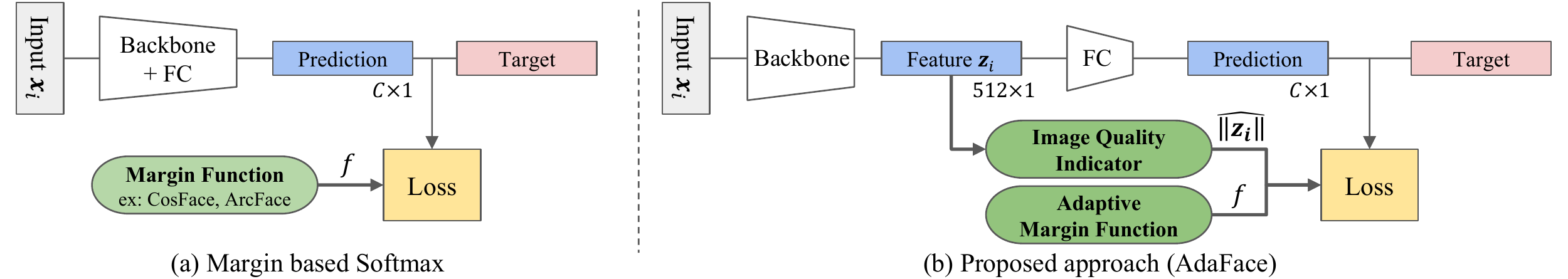}
    \vspace{-5mm}
    \caption{Conventional margin based softmax loss vs our AdaFace. 
    (a)  A FR training pipeline with a margin based softmax loss. 
    The loss function takes the margin function to induce smaller intra-class variations. Some examples are SphereFace, CosFace and ArcFace~\cite{liu2017sphereface, wang2018cosface, deng2019arcface}. 
    (b) Proposed adaptive margin function (AdaFace) that is adjusted based on the image quality indicator. If the image quality is indicated to be low, the loss function emphasizes easy samples (thereby avoiding unidentifiable images). Otherwise, the loss emphasizes hard samples. \vspace{-6mm}}
    \label{figure:adaface_figure2_v2}
\end{figure*}

One problem with low quality face images is that they tend to be unrecognizable. When the image degradation is too large, the relevant identity information vanishes from the image, resulting in \textit{unidentifiable images}. These unidentifiable images are detrimental to the training procedure since a model will try to exploit other visual characteristics, such as clothing color or image resolution, to lower the training loss. If these images are dominant in the distribution of low quality images, the model is likely to perform poorly on low quality datasets during testing.

Motivated by the presence of unidentifiable facial images, we would like to design a loss function which assigns different importance to samples of different difficulties according to the image quality. We aim to emphasize hard samples for the high quality images and easy samples for low quality images. 
Typically, assigning different importance to different difficulties of samples is done by looking at the training progression (curriculum learning)~\cite{bengio2009curriculum, huang2020curricularface}. 
Yet, we show that the sample importance should be adjusted by looking at both the difficulty and the image quality. 

The reason why importance should be set differently according to the image quality is that naively emphasizing hard samples always puts a strong emphasis on unidentifiable images. This is because one can only make a random guess about unidentifiable images and thus, they are always in the hard sample group. There are challenges in introducing image quality into the objective. This is because image quality is a term that is hard to quantify due to its broad definition and scaling samples based on the difficulty often introduces ad-hoc procedures that are heuristic in nature. 

In this work, we present a loss function to achieve the above goal in a seamless way. 
We find that 1) feature norm can be a good proxy for the image quality, and 
2) various margin functions amount to assigning different importance to different difficulties of samples.
These two findings are combined in a unified loss function, AdaFace, that adaptively changes the margin function to assign different importance to different difficulties of samples, based on the image quality (see Fig.~\ref{figure:adaface_figure2_v2}).

In summary, the contributions of this paper include:
 \begin{itemize}
    \item We propose a loss function, AdaFace, that assigns different importance to different difficulties of samples according to their image quality. By incorporating image quality, we avoid emphasizing unidentifiable images while focusing on hard yet recognizable samples.
    \item We show that the angular margin scales the learning signal (gradient) based on the training sample's difficulty. This observation motivates us to change margin function adaptively to emphasize hard samples if the image quality is high, and ignore very hard samples (unidentifiable images) if the image quality is low.
    \item We demonstrate that feature norms can serve as the proxy of image quality. It bypasses the need for an additional module to estimate image quality. Thus, adaptive margin function is achieved without additional complexity. 
    \item We verify the efficacy of the proposed method by extensive evaluations on $9$ datasets (LFW, CFP-FP, CPLFW, AgeDB, CALFW, IJB-B, IJB-C, IJB-S and TinyFace) of various qualities. 
    We show that the recognition performance on low quality datasets can be hugely increased while maintaining performance on high quality datasets. 
\end{itemize}

\section{Related Work}

\Paragraph{Margin Based Loss Function} The margin based softmax loss function is widely used for training face recognition (FR) models~\cite{wang2018cosface, deng2019arcface, liu2017sphereface, huang2020curricularface}. Margin is added to the softmax loss because without the margin, learned features are not sufficiently discriminative. 
SphereFace~\cite{liu2017sphereface}, CosFace~\cite{wang2018cosface} and ArcFace~\cite{deng2019arcface} introduce different forms of margin functions. 
Specifically, it can be written as,
\begin{equation}
\mathcal{L} = - \log \frac{\exp(f(\theta_{y_i}, m))}{\exp(f(\theta_{y_i}, m)) + \sum_{j\neq y_i}^n \exp(s\cos\theta_{j})},
\label{eq:margin_based_softmax}
\end{equation}
where $\theta_j$ is the angle between the feature vector and the $j^{th}$ classifier weight vector, $y_i$ is the index of the ground truth (GT) label, and $m$ is the margin, which is a scalar hyper-parameter. $f$ is a margin function, where
\begin{equation}
    f(\theta_j, m)_{\text{SphereFace}} =
\begin{cases}
    s \cos (m\theta_{j})  & j=y_i \\
    s \cos \theta_{j} & j \neq y_i
\end{cases},
\label{eq:margin_function1}
\end{equation}
\begin{equation}
    f(\theta_j, m)_{\text{CosFace}} = 
\begin{cases}
    s (\cos\theta_{j} - m)  & j=y_i \\
    s \cos \theta_{j} & j \neq y_i
\end{cases},
\label{eq:margin_function2}
\end{equation}
\begin{equation}
    f(\theta_j, m)_{\text{ArcFace}} = 
\begin{cases}
    s \cos (\theta_{j}+m)  & j=y_i \\
    s \cos \theta_{j} & j \neq y_i
\end{cases}.
\label{eq:margin_function3}
\end{equation}
Sometimes, ArcFace is referred to as an \textit{angular} margin and CosFace is referred to as an \textit{additive} margin. Here, $s$ is a hyper-parameter for scaling. P2SGrad~\cite{zhang2019p2sgrad} notes that $m$ and $s$ are sensitive hyper-parameters and proposes to directly modify the gradient to be free of $m$ and $s$.

Our approach aims to model the margin $m$ as a function of the image quality because $f(\theta_{y_i}, m)$ has an impact on which samples contribute more gradient ({\it i.e.} learning signal) during training.

\Paragraph{Adaptive Loss Functions} Many studies have introduced an element of adaptiveness in the training objective for either hard sample mining~\cite{wang2020mis, lin2017focal}, scheduling difficulty during training~\cite{huang2020curricularface, shrivastava2016training}, or finding optimal hyperparameters~\cite{zhang2019adacos}. 
For example, CurricularFace~\cite{huang2020curricularface} 
brings the idea of curriculum learning
into the loss function. During the initial stages of training, the margin for $\cos \theta_j$ (negative cosine similarity) is set to be small so that easy samples can be learned and in the later stages, the margin is increased so that hard samples are learned. Specifically, it is written as 
\begin{equation}
\scriptsize
    f(\theta_j, m)_{\text{Curricular}} = 
\begin{cases}
    s \cos (\theta_{j}+m)  & j=y_i \\
    N(t, \cos \theta_j) & j \neq y_i
\end{cases},
\end{equation}
where 
\begin{equation}
\scriptsize
    N(t, \cos \theta_j) = 
\begin{cases}
    \cos (\theta_{j})  & s \cos (\theta_{y_i}+m)  \ge \cos\theta_j \\
    \cos (\theta_{j})(t+\cos\theta_j) & s \cos (\theta_{y_i}+m)  < \cos\theta_j
\end{cases},
\end{equation}
and $t$ is a parameter that increases as the training progresses. Therefore, in CurricularFace, the adaptiveness in the margin is based on the training progression (curriculum).  

On the contrary, we argue that the adaptiveness in the margin should be based on the image quality. We believe that among high quality images, if a sample is hard (with respect to a model), the network should learn to exploit the information in the image, but in low quality images, if a sample is hard, it is more likely to be devoid of proper identity clues and the network should not try hard to fit on it. 

MagFace~\cite{meng2021magface} explores the idea of applying different margins based on recognizability.
It applies large angular margins to high norm features on the premise that high norm features are easily recognizable. 
Large margin pushes features of high norm closer to class centers. 
Yet, it fails to emphasize hard training samples, which is important for learning discriminative features. A detailed contrast with MagFace can be found in the supplementary B.1. It is also worth mentioning that
DDL~\cite{huang2020improving} uses the distillation loss to minimize the gap between easy and hard sample features.

\Paragraph{Face Recognition with Low Quality Images} Recent FR models have achieved high performance on datasets where facial attributes are discernable,  
{\it e.g.}, LFW\cite{lfw}, CFP-FP\cite{cfpfp}, CPLFW\cite{cplfw}, AgeDB\cite{agedb} and CALFW\cite{calfw}. 
Good performance on these datasets can be achieved when the FR model learns discriminative features invariant to lighting, age or pose variations. 
However, FR in unconstrained scenarios such as in surveillance or low quality videos~\cite{yin2020fan} brings more problems to the table. 
Examples of datasets in this setting are IJB-B\cite{ijbb}, IJB-C\cite{ijbc} and IJB-S\cite{ijbs}, where most of the images are of low quality, and some do not contain sufficient identity information, even for human examiners. 
The key to good performance involves both 1) learning discriminative features for low quality images and 2) learning to discard images that contain few identity cues. 
The latter is sometimes referred to as \textit{quality aware fusion}. 

To perform quality aware fusion, probabilistic approaches have been proposed to predict uncertainty in FR representation~\cite{improving-face-recognition-with-a-quality-based-probabilistic-framework, shi2019probabilistic, chang2020data, li2021spherical,zhou2003probabilistic}. 
They assume the features are distributions and the variance can be used to calculate the certainty in prediction. 
However, probabilistic approaches often resort to learning mean and variance separately, which is not simple during training and suboptimal as the variance is optimized with a fixed mean. Our work, however, is a modification to the conventional softmax loss, making the framework easy to use. 
And we use the feature norm as a proxy for quality during quality-aware fusion.  

QSub-PM~\cite{zheng2020automatic} and UGG~\cite{zheng2019uncertainty} also show good performances in LQ video recognition by using rich subspace (matrix) representation for comparison and using auxiliary context (such as a body) to aid feature fusion respectively.

Synthetic data or augmentations can be used to mimic low quality data. \cite{shi2020towards,feng2018joint} adopts 3D reconstruction to generate faces.
Extra steps complicate the training procedure, making it hard to generalize to other domains.
We adopt easily applicable crop, blur and photometric augmentations.

\section{Proposed Approach}

\begin{figure*}[ht]
    \centering
    \includegraphics[width=\linewidth]{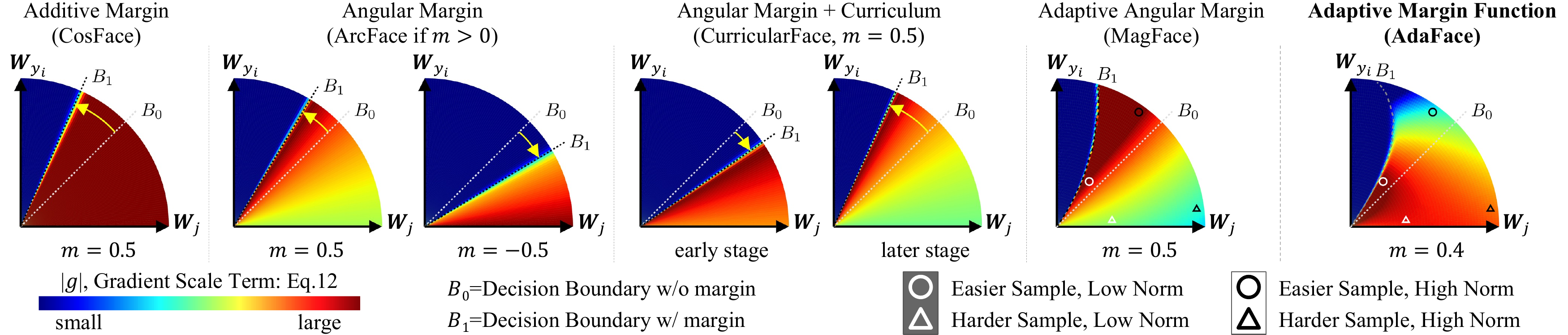}
    \vspace{-5mm}
    \caption{Illustration of different margin functions and their gradient scaling terms on the feature space. $B_0$ and $B_1$ show the decision boundary with and without margin $m$, respectively. 
    The yellow arrow indicates the shift in the boundary due to margin $m$. 
    In the arc, a well-classified sample will be close to (in angle) the ground truth class weight vector, $\bm{W}_{y_i}$. A misclassified sample will be close to $\bm{W}_j$, the negative class weight vector.
    The color within the arc indicates the magnitude of the gradient scaling term $g$ (Eq.~\ref{gradient_scale_def}). Samples in the dark red region will contribute more to learning.
    Note that additive margin shifts the boundary toward $\bm{W}_{y_i}$, without changing the gradient scaling term. 
    However, positive angular margin not only shifts the boundary, but also makes the gradient scale high near the boundary and low away from the boundary. 
    This behavior de-emphasizes very hard samples, and likewise MagFace has similar behavior. On the other hand, negative angular margin induces an opposite behavior. 
    CurricularFace adapts the boundary based on the training stage. 
    Our work adaptively changes the margin functions based on the norm. With high norm, we emphasize samples away from the boundary and with low norm we emphasize samples near the boundary. 
    Circles and triangles in the arc show example scenarios in the right most plot (AdaFace). \vspace{-3mm}}
    \label{figure:architecture}
\end{figure*}


The cross entropy softmax loss of a sample $\bm{x}_i$ can be formulated as follows,
\begin{equation}
\mathcal{L}_{CE}(\bm{x}_i) = - \log \frac{\exp(\bm{W}_{y_i} \bm{z}_i + b_{y_i})}{ \sum_{j=1}^C \exp(\bm{W}_{j} \bm{z}_j + b_{j})},
\label{eq:softmax}
\end{equation}
where $\bm{z}_i \in \mathbb{R}^d$ is the $\bm{x}_i$'s feature embedding, and $\bm{x}_i$ belongs to the $y_i$th class. $\bm{W}_j$ refers to the $j$th column of the last FC layer weight matrix, $\bm{W} \!\in \mathbb{R}^{d \times C }$, and $b_j$ refers to the corresponding bias term. $C$ refers to the number of classes. 

During test time, for an arbitrary pair of images, $\bm{x}_p$ and $\bm{x}_q$, the cosine similarity metric, $\frac{\bm{z}_p \cdot \bm{z}_q }{\Vert \bm{z}_p \Vert \Vert \bm{z}_q \Vert}$ is used to find the closest matching identities. To make the training objective directly optimize the cosine distance, \cite{liu2017sphereface, wang2017normface} use normalized softmax where the bias term is set to zero and the feature $\bm{z}_i$ is normalized and rescaled with $s$ during training. This modification results in 
\begin{equation}
\mathcal{L}_{CE}(\bm{x}_i) = - \log \frac{\exp(s \cdot \cos\theta_{y_i})}{ \sum_{j=1}^C \exp(s\cos\theta_{j})},
\label{eq:cos_softmax}
\end{equation}
where $\theta_j$ corresponds to the angle between $\bm{z}_i$ and $\bm{W}_j$. Follow-up works~\cite{deng2019arcface, wang2018cosface} take this formulation and introduces a margin to reduce the intra-class variations. Generally, it can be written as Eq.~\ref{eq:margin_based_softmax}
where margin functions are defined in Eqs.~\ref{eq:margin_function1}, \ref{eq:margin_function2} and \ref{eq:margin_function3} correspondingly.

\subsection{Margin Form and the Gradient}
\label{sec:margin_form}
Previous works on margin based softmax focused on how the margin shifts the decision boundaries and what their geometric interpretations are~\cite{wang2018cosface, deng2019arcface}. In this section, we show that during backpropagation, the gradient change due to the margin has the effect of scaling the importance of a sample relative to the others. In other words, angular margin can introduce an additional term in the gradient equation that scales the signal according to the sample's difficulty. To show this, we will look at how the gradient equation changes with the margin function $f(\theta_{y_i}, m)$.

Let $P_j^{(i)}$ be the probability output at class $j$ after softmax operation on an input $\bm{x}_i$. By deriving the gradient equations for $\mathcal{L}_{CE}$ w.r.t.~$\bm{W}_j$ and $\bm{x}_i$, we obtain the following,
\begin{equation}
    P_j^{(i)} = \frac{\exp(f(\cos \theta_{y_i}) )}{\exp(f(\cos \theta_{y_i})) + \sum_{j\neq y_i}^n \exp(s\cos\theta_{j})},
\end{equation}
\begin{equation}
    \frac{\partial \mathcal{L_{\text{CE}}} }{ \partial \bm{W}_j} = \left( P_j^{(i)} - \mathbbm{1}(y_i = j) \right) \frac{\partial f(\cos \theta_j) }{ \partial \cos \theta_j} \frac{ \partial \cos \theta_j}{ \partial \bm{W}_j},
    \label{partialw}
\end{equation}
\begin{equation}
    \frac{\partial \mathcal{L_{\text{CE}}}}{ \partial\bm{x}_i} \!=\! \sum_{k=1}^C \!\left( P_k^{(i)} \!-\! \mathbbm{1}(y_i = k) \!\right)\! \frac{\partial f(\cos \theta_k) }{ \partial \cos \theta_k}\! \frac{ \partial \cos \theta_k}{ \partial \bm{x}_i}.
    \label{partialx}
\end{equation}
In Eqs.~\ref{partialw} and \ref{partialx}, the first two terms, $\left( P_j^{(i)} - \mathbbm{1}(y_i = j) \right)$ and $\frac{\partial f(\cos \theta_j) }{ \partial \cos \theta_j}$ are scalars. 
Also, these two are the only terms affected by parameter $m$ through $f(\cos \theta_{y_i})$. 
As the direction term, $\frac{ \partial \cos \theta_j}{ \partial \bm{W}_j}$ is free of $m$, we can think of the first two scalar terms as a gradient scaling term (GST) and denote, 
\begin{equation}
    g := \left( P_j^{(i)} - \mathbbm{1}(y_i = j) \right) \frac{\partial f(\cos \theta_j) }{ \partial \cos \theta_j}.
    \label{gradient_scale_def}
\end{equation}

For the purpose of the GST analysis, we will consider the class index $j=y_i$, since all negative class indices $j\neq y_i$ do not have a margin in  Eqs.~\ref{eq:margin_function1}, \ref{eq:margin_function2}, and \ref{eq:margin_function3}. The GST for the normalized softmax loss is 
\begin{equation}
    g_{\text{softmax}} = (P_{y_i}^{(i)}-1) s,
\end{equation}
since $f(\cos\theta_{y_i}) = s\cdot \cos \theta_{y_i} $ and  $\frac{\partial f(\cos \theta_{y_i}) }{ \partial \cos \theta_{y_i}} = s$. The GST for the CosFace\cite{wang2018cosface} is also
\begin{equation}
    g_{\text{CosFace}} = (P_{y_i}^{(i)}-1) s,
\end{equation}
as $f(\cos\theta_{y_i}) = s (\cos \theta_{y_i} - m)$ and $\frac{\partial f(\cos \theta_{y_i}) }{ \partial \cos \theta_{y_i}} = s$. 
Yet, the GST for ArcFace\cite{deng2019arcface} turns out to be
\begin{equation}
    g_{\text{ArcFace}} \!= (P_j^{(i)}\!-1) s \left( \cos(m) \!+ \frac{\cos \theta_{y_i} \sin(m)}{\sqrt{1 \!- \cos^2 \theta_{y_i}}} \right).
    \label{g_arcface}
\end{equation}
The derivation can be found in the supplementary. 
Since the GST is a function of $\theta_{y_i}$ and $m$ as in Eq.~\ref{g_arcface}, it is possible to use it to control the emphasis on samples based on the difficulty, {\it i.e.}, $\theta_{y_i}$ during training. 

To understand the effect of GST, we visualize GST \textit{w.r.t}.~the features. 
Fig.~\ref{figure:architecture} shows the GST as the color in the feature space. 
Note that for the angular margin, the GST peaks at the decision boundary but slowly decreases as it moves away towards $\bm{W}_j$ and harder samples receive less emphasis. 
If we change the sign of the angular margin, we see an opposite effect. Note that, in the $6$th column, MagFace~\cite{meng2021magface} is an extension of ArcFace (positive angular margin) with larger margin assigned to high norm feature. 
Both ArcFace and MagFace fail to put high emphasis on hard samples (green area near $\bm{W}_{j}$). We combine all margin functions (positive and negative angular margins and additive margins) to emphasize hard samples when necessary.

Note that this adaptiveness is also different from approaches that use the training stage to change the relative importance of different difficulties of samples~\cite{huang2020curricularface}. Fig.~\ref{figure:architecture} shows CurricularFace where the decision boundary and the GST $g$ change depending on the training stage. 



\subsection{Norm and Image quality}
\label{sec:norm_image_quality}
Image quality is a comprehensive term that covers characteristics such as brightness, contrast and sharpness.  
Image quality assessment (IQA) is widely studied in computer vision~\cite{zhai2020perceptual}. SER-FIQ~\cite{serfiq} is an unsupervised DL method for face IQA.
BRISQUE~\cite{brisque} is a popular algorithm for blind/no-reference IQA. 
However, such methods are computationally expensive to use during training.  
In this work, we refrain from introducing an additional module that calculates the image quality. 
Instead, we use the feature norm as a proxy for the image quality. 
We observe that, in models trained with a margin-based softmax loss, the feature norm exhibits a trend that is correlated with the image quality. 

\begin{figure}[t]
\centering
  \includegraphics[width=\linewidth]{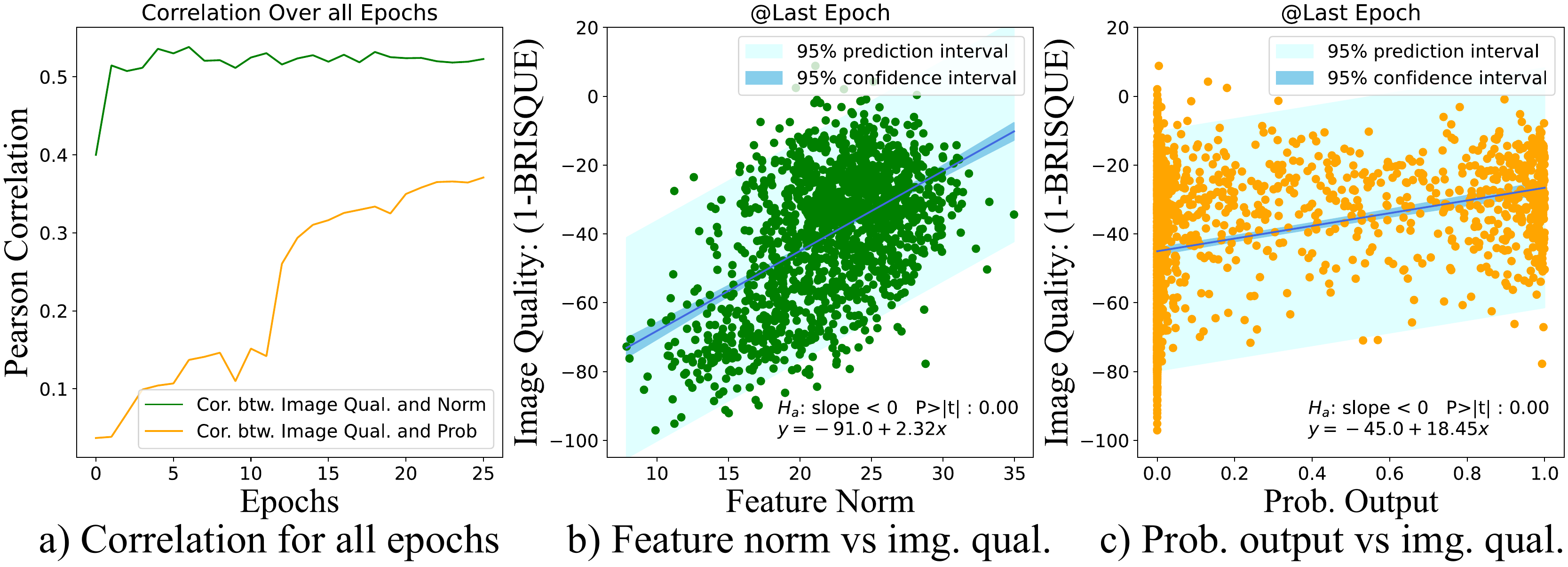}
  \vspace{-5mm}
  \caption{(a) A plot of Pearson correlation with image quality score (1-BRISQUE) over training epochs. The green and orange curves correspond to the correlation plot using the feature norm $\Vert \mathbf{z}_i \Vert$ and the probability output for the ground truth index $P_{y_i}$, respectively. (b) and (c) Corresponding scatter plots for the last epoch. The blue line on the scatter plot and the corresponding equation shows the least square line fitted to the data points.}
  \label{figure:norm_plot}
  \vspace{-2mm}
\end{figure}

In Fig.~\ref{figure:norm_plot} (a) we show a correlation plot between the feature norm and the image quality (IQ) score calculated with ($1$-BRISQUE) as a green curve. We randomly sampled $1,534$ images from the training dataset (MS1MV2~\cite{deng2019arcface} with augmentations described in Sec.~\ref{implementation}) and calculate the feature norm using a pretrained model. At the final epoch, the correlation score between the feature norm and IQ score reaches  $0.5235$ (out of $-1$ and $1$). The corresponding scatter plot is shown in Fig.~\ref{figure:norm_plot} (b). This high correlation between the feature norm and the IQ score supports our use of feature norm as the proxy of image quality. 

In Fig.~\ref{figure:norm_plot} (a) we also show a correlation plot between the probability output $P_{y_i}$ and the IQ score as an orange curve. Note that the correlation is always higher for the feature norm than for $P_{y_i}$. Furthermore, the correlation between the feature norm and IQ score is visible from an early stage of training. This is a useful property for using the feature norm as the proxy of image quality because we can rely on the proxy from the early stage of training. Also, in Fig.~\ref{figure:norm_plot} (c), we show a scatter plot between $P_{y_i}$ and IQ score. Notice that there is a non-linear relationship between $P_{y_i}$ and the image quality. One way to describe a sample's difficulty is with $1-P_{y_i}$, and the plot shows that the distribution of the difficulty of samples is different based on image quality. Therefore, it makes sense to consider the image quality when adjusting the sample importance according to the difficulty.

\subsection{AdaFace: Adaptive Margin based on Norm}
\label{adaface_defined}
To address the problem caused by the unidentifiable images, we propose to adapt the margin function based on the feature norm. 
In Sec.~\ref{sec:margin_form}, we have shown that using different margin functions can emphasize different difficulties of samples. 
Also, in Sec.~\ref{sec:norm_image_quality}, we have observed that the feature norm can be a good way to find low quality images. 
We will merge the two findings and propose a new loss for FR.

\Paragraph{Image Quality Indicator}
\label{image_quality_approx}
\newcommand{\clip}[1]{\left\lfloor#1\right\rceil}
As the feature norm, $\Vert \bm{z}_i\Vert$ is a model dependent quantity, we normalize it using batch statistics $\mu_z$ and $\sigma_z$. Specifically, we let
\begin{equation}
    \widehat{\Vert \bm{z}_i \Vert} = \clip{\frac{\Vert \bm{z}_i \Vert - \mu_{z}}{\sigma_{z} / h}}^1_{-1},
\end{equation}
where $\mu_z$ and $\sigma_z$  are the mean and standard deviation of all $\Vert \bm{z}_i \Vert$ within a batch. And $\clip{\cdot}$ refers to clipping the value between $-1$ and $1$ and stopping the gradient from flowing. 
Since $\frac{\Vert \bm{z}_i \Vert - \mu_{z}}{\sigma_{z} / h}$ makes the batch distribution of $\widehat{\Vert \bm{z}_i \Vert}$ as approximately unit Gaussian, we clip the value to be within $-1$ and $1$ for better handling. 
It is known that approximately $68\%$ of the unit Gaussian distribution falls between $-1$ and $1$, so we introduce the term $h$ to control the concentration. 
We set $h$ such that most of the values $\frac{\Vert \bm{z}_i \Vert - \mu_{z}}{\sigma_{z} / h}$ fall between $-1$ and $1$. 
A good value to achieve this would be $h=0.33$. 
Later in 
Sec.~\ref{ablation}, we ablate and validate this claim. We stop the gradient from flowing during backpropagation because we do not want features to be optimized to have low norms.

If the batch size is small, the batch statistics $\mu_z$ and $\sigma_z$ can be unstable. 
Thus we use the exponential moving average (EMA) of $\mu_z$ and $\sigma_z$ across multiple steps to stabilize the batch statistics. 
Specifically, let $\mu^{(k)}$ and $\sigma^{(k)}$ be the $k$-th step batch statistics of $\Vert \bm{z}_i \Vert$. Then 
\begin{equation}
    \mu_z = \alpha \mu_z^{(k)} + (1-\alpha) \mu_z^{(k-1)}, 
\end{equation} 
and $\alpha$ is a momentum set to $0.99$. The same is true for $\sigma_z$.

\Paragraph{Adaptive Margin Function}
We design a margin function such that 1) if image quality is high, we emphasize hard samples, and 2) if image quality is low, we de-emphasize hard samples. 
We achieve this with two adaptive terms $g_\text{angle}$ and $g_\text{add}$, referring to angular and additive margins, respectively. Specifically, we let
\begin{equation}
    f(\theta_j, m)_{\text{AdaFace}}\! = \!
\begin{cases}
    s (\cos (\theta_{j} \!+\! g_{\text{angle}})\!-\!g_{\text{add}}) & j\!=\!y_i \\
    s \cos \theta_{j} & j \!\neq \!y_i
\end{cases}
\end{equation} 
where $g_{\text{angle}}$ and $g_{\text{add}}$ are the functions of $\widehat{\Vert \bm{z}_i \Vert}$. We define 
\begin{equation}
    g_{\text{angle}} = -m \cdot \widehat{\Vert \bm{z}_i \Vert},\quad
    g_{\text{add}} = m \cdot \widehat{\Vert \bm{z}_i \Vert} + m.
\end{equation}
Note that when $\widehat{\Vert \bm{z}_i \Vert} = -1$, the proposed function becomes ArcFace. When $\widehat{\Vert \bm{z}_i \Vert} = 0$, it becomes CosFace. When $\widehat{\Vert \bm{z}_i \Vert} = 1$, it becomes a negative angular margin with a shift. 
Fig.~\ref{figure:architecture} shows the effect of the adaptive function on the gradient. 
The high norm features will receive a higher gradient scale, far away from the decision boundary, whereas
the low norm features will receive higher gradient scale near the decision boundary. 
For low norm features, the harder samples away from the boundary are de-emphasized. 


\section{Experiments}\label{sec:experiment}

\subsection{Datasets and Implementation Details}
\label{implementation}
\begin{figure}[t!]
\centering
  \includegraphics[width=0.97\linewidth]{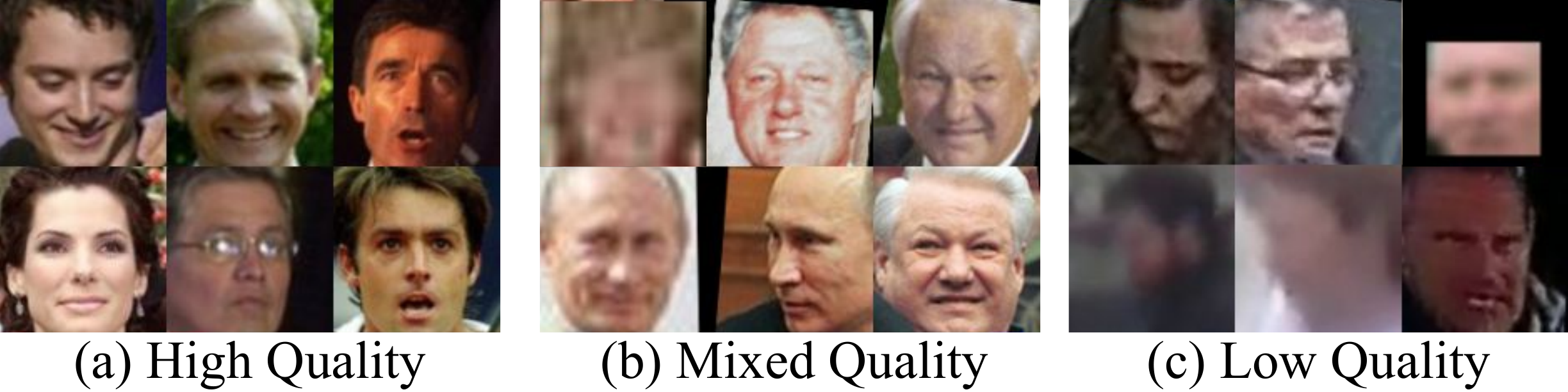}
  \vspace{-1.5mm}
  \caption{Examples of three categories of test datasets in our study.\vspace{-4mm}}
  \label{figure:img_example}
\end{figure}

\Paragraph{Datasets} We use MS1MV2\cite{deng2019arcface}, MS1MV3~\cite{deng2019lightweight} and WebFace4M~\cite{zhu2021webface260m} as our training datasets. Each dataset contains $5.8$M, $5.1$M and $4.2$M facial images, respectively. 
We test on $9$ datasets of varying qualities. 
Following the protocol of~\cite{shi2020towards}, we categorize the test datasets into $3$ types according to the visual quality (examples shown in Fig.~\ref{figure:img_example}).
\begin{itemize}
    \item \textbf{High Quality}: LFW\cite{lfw}, CFP-FP\cite{cfpfp}, CPLFW\cite{cplfw} AgeDB\cite{agedb} and CALFW\cite{calfw} are popular benchmarks for FR in the well controlled setting. While the images show variations in lighting, pose, or age, they are of sufficiently good quality for face recognition. 
    \item \textbf{Mixed Quality}: IJB-B and IJB-C~\cite{ijbb, ijbc} are datasets collected for the purpose of introducing low quality images in the validation protocol. They contain both high quality images and low quality videos of celebrities.
    \item \textbf{Low Quality}: IJB-S~\cite{ijbs} and TinyFace~\cite{tinyface} are datasets with low quality images and/or videos. 
    IJB-S is a surveillance video dataset, with test protocols such as 
    \textit{Surveillance-to-Single, Surveillance-to-Booking and Surveillance-to-Surveillance}. 
    The first/second word in the protocol refers to the probe/gallery image source. 
    \textit{Surveillance} refers to the surveillance video, \textit{Single} refers to a high quality enrollment image and \textit{Booking} refers to multiple enrollment images taken from different viewpoints. TinyFace consists only of low quality images.

\end{itemize}


\Paragraph{Training Settings}
We preprocess the dataset by cropping and aligning faces with five landmarks, as in~\cite{zhang2016joint, deng2019arcface}, resulting in  $112 \times 112$ images. For the backbone, we adopt ResNet~\cite{he2016deep} as modified in~\cite{deng2019arcface}. 
We use the same optimizer and a learning rate schedule as in~\cite{huang2020curricularface}, and train for $24$ epochs. 
The model is trained with SGD with the initial learning rate of 0.1 and step scheduling at $10$, $18$ and $22$ epochs. If the dataset contains augmentations, we add $2$ more epochs for convergence. For the scale parameter $s$, we set it to $64$, following the suggestion of \cite{deng2019arcface, wang2018cosface}.

\Paragraph{Augmentations}
Since our proposed method is designed to train better in the presence of unidentifiable images in the training data, we introduce three on-the-fly augmentations that are widely used in image classification tasks~\cite{he2019bag}, \textit{i.e.}, cropping, rescaling and photometric jittering. These augmentations will create more data but also introduce more unidentifiable images. It is a trade-off that has to be balanced. In FR, these augmentations are not used because they generally do not bring benefit to the performance (as shown in Sec.~\ref{ablation}). We show that our loss function is capable of reaping the benefit of augmentations because it can adapt to ignore unidentifiable images.   

Cropping defines a random rectangular area (patch) and makes the region outside the area to be $0$. We do not cut and resize the image as the alignment of the face is important. 
Photometric augmentation randomly scales hue, saturation and brightness. Rescaling involves resizing an image to a smaller scale and back, resulting in blurriness. 
These operations are applied randomly with a probability of $0.2$. 

\subsection{Ablation and Analysis}
\label{ablation}
For hyperparameter $m$ and $h$ ablation, we adopt  a ResNet18 backbone and use $1/6$th of the randomly sampled MS1MV2. 
We use two performance metrics. 
For High Quality Datasets (HQ), we use an average of $1$:$1$ verification accuracy in
LFW, CFP-FP, CPLFW, AgeDB and CALFW. 
For Low Quality Datasets (LQ), we use an average of the closed-set rank-$1$ retrieval and the open-set TPIR@FIPR=$1\%$  for all $3$ protocols of IJB-S. 
Unless otherwise stated, we augment the data as described in Sec.~\ref{implementation}.

\Paragraph{Effect of Image Quality Indicator Concentration $h$} 
In Sec.~\ref{image_quality_approx}, we claim that $h=0.33$ is a good value.
To validate this claim, we show in Tab.~\ref{table:ablation_hyper} the performance when varying $h$. 
When $h=0.33$, the model performs the best. For $h=0.22$ or $h=0.66$, the performance is still higher than CurricularFace. As long as $h$ is set such that $\widehat{\Vert \bm{z}_i \Vert}$ has some variation, $h$ is not very sensitive. We set $h=0.33$. 

\Paragraph{Effect of Hyperparameter $m$}
The margin $m$ corresponds to both the maximum range of the angular margin and the magnitude of the additive margin. 
Tab.~\ref{table:ablation_hyper} shows that the performance is best for HQ datasets when $m=0.4$ and for LQ datasets when $m=0.75$. Large $m$ results in large angular margin variation based on the image quality, resulting in more adaptivity. In subsequent experiments, we choose $m=0.4$ since it achieves good performance for LQ datasets without sacrificing performance on HQ datasets.

\Paragraph{Effect of Proxy Choice} In Tab.~\ref{table:ablation_hyper}, to show the effectiveness of using the feature norm as a proxy for image quality, we switch the feature norm with other quantities such as (1-BRISQUE) or $P_{y_i}$. The performance using the feature norm is superior to using others. The BRISQUE score is precomputed for the training dataset, so it is not as effective in capturing the image quality when training with augmentation. We include $P_{y_i}$ to show that the adaptiveness in feature norm is different from adaptiveness in difficulty.

\begin{table}[!t]
\centering
\scriptsize
\setlength{\tabcolsep}{3pt}
\begin{tabular}{|c|c|c|c||c|c|}
\hline
Method & $h$           & $m$             & Proxy    & HQ Datasets & LQ Datasets   \\ \hline\hline
CurricularFace~\cite{huang2020curricularface} &    -     & $0.50$   &   &    $93.43$   &   $32.92$           \\ \hline
\textcolor{white}{aaa}AdaFace\textcolor{white}{aaa} & $0.22$         &  \multirow{3}{*}{$0.40$}    & \multirow{3}{*}{Norm}    &   $93.67$    &     $34.92$          \\
AdaFace & $\bm{0.33}$        &    &     &    $\bm{93.74}$   &    $\bm{35.40}$           \\
AdaFace & $0.66$        &      &     &    $93.70$   &  $35.29$    \\ \hline\hline
\textcolor{white}{aaa}AdaFace\textcolor{white}{aaa} & \multirow{3}{*}{$0.33$}         &    $\bm{0.40}$  & \multirow{3}{*}{Norm} &   $\bm{93.74}$    &  $35.40$           \\
AdaFace &         &    $0.50$ &   &  $93.56$     &      $35.23$      \\
AdaFace &  &    $0.75$ &  &    $93.37$   &    $\bm{35.69}$             \\ \hline\hline
\textcolor{white}{aaa}AdaFace\textcolor{white}{aaa} & \multirow{3}{*}{$0.33$}         &    \multirow{3}{*}{$0.40$}   & \textbf{Norm} & $\bm{93.74}$    &  $\bm{35.40}$           \\
- &    &     & $1-$BRISQUE   &  $93.43$     &      $34.55$      \\
- &  &  & $P_{y_i}$ &  $93.46$   &    $35.17$             \\ \hline
\end{tabular}
\vspace{-2mm}
\caption{Ablation of our margin function parameters $h$ and $m$, and the image quality proxy choice on the ResNet18 backbone. The performance metrics are as described in Sec.~\ref{ablation}. \vspace{-3mm}}
\label{table:ablation_hyper}
\end{table}

\begin{table}[t]
\centering
\scriptsize
\setlength{\tabcolsep}{3pt}
\begin{tabular}{|c|c||c|c|}
\hline
Method & $p$   & HQ Datasets & LQ Datasets   \\ \hline\hline
CurricularFace~\cite{huang2020curricularface}    &  $\bm{0.0}$    &   $\bm{96.85}$    &   $\bm{41.00}$             \\
CurricularFace~\cite{huang2020curricularface}  &    $0.2$   &    $96.75$   &     $40.84$           \\
CurricularFace~\cite{huang2020curricularface}   &     $0.3$  &    $96.59$   &      $40.58$          \\ \hline
AdaFace    &  $0.0$    &   $96.72$    &   $40.95$            \\
AdaFace  &    $\bm{0.2}$   &    $\bm{96.88}$   &    $41.82$         \\
AdaFace   &     $0.3$  &    $96.78$   &    $\bm{41.93}$          \\ \hline
\end{tabular}
\vspace{-2mm}
\caption{Ablation of augmentation probability $p$, on the ResNet50 backbone. The metrics are the same as Tab.~\ref{table:ablation_hyper}. \vspace{-3mm}}
\label{table:ablation_aug}
\end{table}

\Paragraph{Effect of Augmentation} We introduce on-the-fly augmentations in our training data. Our proposed loss can effectively handle the unidentifiable images, which are generated occasionally during augmentations.
We experiment with a larger model ResNet50 on the full MS1MV2 dataset. 

Tab.~\ref{table:ablation_aug} shows that indeed the augmentation brings performance gains for AdaFace. The performance on HQ datasets stays the same, whereas LQ datasets enjoy a significant performance gain. Note that the augmentation hurts the performance of CurricularFace, which is in line with our assumption that augmentation is a tradeoff between a positive effect from getting more data and a negative effect from unidentifiable images. Prior works on margin-based softmax do not include on-the-fly augmentations as the performance could be worse. AdaFace avoids overfitting on unidentifiable images, therefore it can exploit the augmentation better. 

\begin{figure}[t]
\centering
  \includegraphics[width=\linewidth]{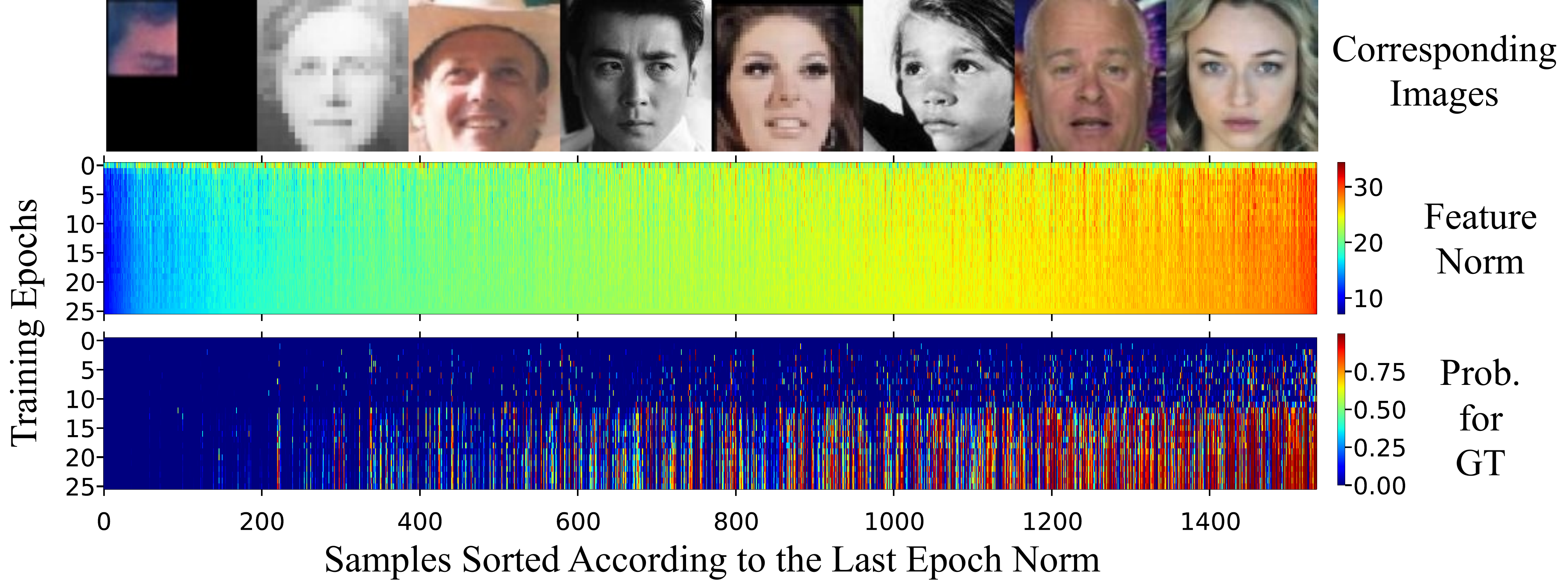}
  \vspace{-6mm}
  \caption{A plot of training samples' trajectories of feature norm $\Vert \bm{z}_i\Vert$ and the probability output for the ground truth index $P_{y_i}$. 
  We randomly select $1,536$ samples from the training data with augmentations, and show $8$ images evenly sampled from them.
  The features with low norm have a different probability trajectory than others and the corresponding images are hard to identify. \vspace{-3mm}}
  \label{figure:norm_prob_plot}
\end{figure}

\Paragraph{Analysis}
To show how the feature norm $\Vert \bm{z}_i\Vert$ and the difficulty of training samples change during training, we plot the sample trajectory in Fig.~\ref{figure:norm_prob_plot}. A total of $1,536$ samples are randomly sampled from the training data. Each column in the heatmap represents a sample, and the x-axis is sorted according to the norm of the last epoch. Sample \#$600$ is approximately a middle point of the transition from low to high norm samples. The bottom plot shows that many of the probability trajectories of low norm samples never get high probability till the end. 
It is in line with our claim that low norm features are more likely to be unidentifiable images. It justifies our motivation to put less emphasis on these cases, although they are “hard” cases. The percentage of samples with augmentations is higher for the low norm features than for the high norm features. For samples number \#$0$ to \#$600$, about $62.0$\% are with at least one type of augmentation. For the samples \#$600$ or higher, the percentage is about $38.5$\%. 

\Paragraph{Time Complexity}
Compared to classic margin-based loss functions, our method adds a negligible amount of computation in training. With the same setting, ArcFace~\cite{deng2019arcface} takes $0.3193$s per iteration while AdaFace takes $0.3229$s (+1\%).

\subsection{Comparison with SoTA methods}

\setlength\dashlinedash{0.3pt}
\setlength\dashlinegap{1.5pt}
\setlength\arrayrulewidth{0.4pt}

\begin{table*}[!h]
\centering
\scriptsize
\begin{tabular}{|l|c|c|ccccc|c|cc|}
\hline
\multicolumn{1}{|c|}{\multirow{2}{*}{Method}} & \multicolumn{1}{c|}{\multirow{2}{*}{Venue}} & \multicolumn{1}{c|}{\multirow{2}{*}{Train Data}} &\multicolumn{6}{c|}{High Quality}                                & \multicolumn{2}{c|}{Mixed Quality}               \\ \cline{4-11} 
\multicolumn{1}{|c|}{}     &         &     & \multicolumn{1}{l|}{\scalebox{0.85}{LFW~\cite{lfw}}}   & \multicolumn{1}{l|}{\scalebox{0.85}{CFP-FP~\cite{cfpfp}}} & \multicolumn{1}{l|}{\scalebox{0.85}{CPLFW~\cite{cplfw}}} & \multicolumn{1}{l|}{\scalebox{0.85}{AgeDB~\cite{agedb}}} & \multicolumn{1}{l|}{\scalebox{0.85}{CALFW~\cite{calfw}}} & \multicolumn{1}{c|}{AVG} & \multicolumn{1}{c|}{IJB-B~\cite{ijbb}} & \multicolumn{1}{c|}{IJB-C~\cite{ijbc}} \\ \hline
CosFace ($m=0.35$)~\cite{wang2018cosface} & CVPR18          &  MS1MV2         & $99.81$                   & $98.12$                      & $92.28$                     & $98.11$                     & $95.76$                     & $96.82$                   & $94.80$                     & $96.37$                     \\
ArcFace ($m=0.50$)~\cite{deng2019arcface} & CVPR19       &     MS1MV2          & $\second{99.83}$                   & $98.27$                      & $92.08$                     & $98.28$                     & $95.45$                     & $96.78$                   & $94.25$                     & $96.03$                     \\
AFRN~\cite{kang2019attentional} & ICCV19               &      MS1MV2          & $\first{99.85}$                   & $95.56$                      & $\second{93.48}$                     & $95.35$                     & $\first{96.30}$                     & $96.11$                   & $88.50$                     & $93.00$                     \\
MV-Softmax~\cite{wang2020mis} & AAAI20           &  MS1MV2       & $99.80$                   & $98.28$                      & $92.83$                     & $97.95$                     & $96.10$                     & $96.99$                   & $93.60$                     & $95.20$                     \\
CurricularFace~\cite{huang2020curricularface} & CVPR20       &   MS1MV2           & $99.80$                   & $98.37$                      & $93.13$                     & $\second{98.32}$                     & $\second{96.20}$                     & $97.16$                   & $94.80$                     & $96.10$                     \\
URL~\cite{shi2020towards} &CVPR20        & MS1MV2 & $99.78$ & $\first{98.64}$ & - & - & - & - & - & $\second{96.60}$ \\
BroadFace~\cite{kim2020broadface} & ECCV20           &  MS1MV2 & $\first{99.85}$ & $\second{98.63}$ & $93.17$ & $\first{98.38}$ & $\second{96.20}$ & $\first{97.25}$ & $94.97$ & $96.38$ \\

MagFace~\cite{meng2021magface} & CVPR21         &      MS1MV2         & $\second{99.83}$ & $98.46$ & $92.87$ & $98.17$ & $96.15$ & $97.10$ & $94.51$ & $95.97$                    \\ 
SCF-ArcFace~\cite{li2021spherical} & CVPR21         &      MS1MV2         & $99.82$                   & $98.40$                      & $93.16$                     & $98.30$                     & $96.12$                     & $97.16$                   & $94.74$                     & $96.09$                    \\ 
DAM-CurricularFace~\cite{liu2021dam} & ICCV21         &      MS1MV2         & -                  & -                    & -           & -            & -                 & -          & $\second{95.12}$             & $96.20$                  \\ \hdashline
\textbf{AdaFace ($m=0.4$)}      &   CVPR22    &       MS1MV2              & $99.82$                   & $98.49$                      & $\first{93.53}$                     & $98.05$                     & $96.08$                     & $\second{97.19}$                 & $\first{95.67}$                     & $\first{96.89}$                    
                    \\ \hline \hline
VPL-ArcFace~\cite{deng2021variational} & CVPR21 & MS1MV3 & $\first{99.83}$ & $\first{99.11}$ & $93.45$ & $\first{98.60}$ & $\first{96.12}$ & $\first{97.42}$ & $95.56$ & $96.76$               \\ \hdashline
\textbf{AdaFace ($m=0.4$)}      &  CVPR22 &     MS1MV3          & $\first{99.83}$ & $99.03$ & $\first{93.93}$ & $98.17$ & $96.02$ & $97.40$ & $\first{95.84}$ & $\first{97.09}$      \\ \hline \hline
ArcFace*~\cite{deng2019arcface}& CVPR19 & \scalebox{0.85}{WebFace4M} & $\first{99.83}$ & $\first{99.19}$ & $94.35$ & $\first{97.95}$ & $96.00$ & $97.46$ & $95.75$ & $97.16$                  \\ \hdashline
\textbf{AdaFace ($m=0.4$)}     & CVPR22 &     \scalebox{0.85}{WebFace4M}  & $99.80$ & $99.17$ & $\first{94.63}$ & $97.90$ & $\first{96.05}$ & $\first{97.51}$ & $\first{96.03}$ & $\first{97.39}$           \\ \hline
\end{tabular}
\small
\\(a) A performance comparison of recent methods on high and mixed quality datasets. 
\scriptsize
\\\textcolor{white}{a}\\
\scalebox{0.993}{
\begin{tabular}{|l|c|ccc|ccc|ccc|cc|}
\hline
\multicolumn{1}{|c|}{\multirow{3}{*}{Method}} & \multicolumn{1}{c|}{\multirow{3}{*}{Train Data}}  &\multicolumn{11}{c|}{Low Quality (IJB-S~\cite{ijbs} and TinyFace~\cite{tinyface})}  \\ 
\cline{3-13}   &    & \multicolumn{3}{c|}{Surveillance-to-Single~\cite{ijbs}}     & \multicolumn{3}{c|}{Surveillance-to-Booking~\cite{ijbs}}    & \multicolumn{3}{c|}{Surveillance-to-Surveillance~\cite{ijbs}}  & \multicolumn{2}{c|}{TinyFace~\cite{tinyface}}   \\
           &      & \multicolumn{1}{c}{Rank-$1$} & \multicolumn{1}{c}{Rank-$5$} & \multicolumn{1}{c|}{$1\%$} & \multicolumn{1}{c}{Rank-$1$} & \multicolumn{1}{c}{Rank-$5$} & \multicolumn{1}{c|}{$1\%$} & \multicolumn{1}{c}{Rank-$1$} & \multicolumn{1}{c}{Rank-$5$} & \multicolumn{1}{c|}{$1\%$} & \multicolumn{1}{c}{Rank-$1$} & \multicolumn{1}{c|}{Rank-$5$} \\ \hline
PFE~\cite{shi2019probabilistic}\scalebox{0.875}{\textcolor{white}{aaa}}      & MS1MV2~\cite{deng2019arcface}            & $50.16$         & $58.33$      & $31.88$       & $53.60$          & $61.75$    & $35.99$       & $9.20$           & $20.82$   & $0.84$  & -  & -    \\
ArcFace~\cite{deng2019arcface}      & MS1MV2~\cite{deng2019arcface}            & $57.35$         & $64.42$      & $41.85$       & $57.36$         & $64.95$    & $41.23$       & -             & -       & -     & - & -   \\
URL~\cite{shi2020towards}            & MS1MV2~\cite{deng2019arcface}            & $59.79$         & $65.78$      & $41.06$       & $61.98$         & $67.12$    & $42.73$       & -             & -       & -    & $\second{63.89}$ &  $\second{68.67}$   \\
CurricularFace*~\cite{huang2020curricularface}  & MS1MV2~\cite{deng2019arcface}            & $\second{62.43}$         & $\second{68.68}$      & $\second{47.68}$       & $\second{63.81}$         & $\second{69.74}$    & $\second{47.57}$       & $\second{19.54}$         & $\second{32.80}$   & $\first{2.53}$     & $63.68$ & $67.65$ \\ \hdashline
\textbf{AdaFace} ($m=0.4$)            & MS1MV2~\cite{deng2019arcface}            & $\first{65.26}$         & $\first{70.53}$      & $\first{51.66}$       & $\first{66.27}$         & $\first{71.61}$    & $\first{50.87}$       & $\first{23.74}$         & $\first{37.47}$   & $\second{2.50}$    &  $\first{68.21}$ & $\first{71.54}$   \\ \hline \hline
\textbf{AdaFace ($m=0.4$)}       & MS1MV3~\cite{deng2019lightweight}     & $67.12$         & $72.67$      & $53.67$       & $67.83$         & $72.88$    & $52.03$       & $26.23$         & $40.60$   & $3.28$   & $67.81$  & $70.98$ \\\hline \hline
ArcFace*~\cite{deng2019arcface}         & \scalebox{0.86}{WebFace4M~\cite{zhu2021webface260m}}     & $69.26$ & $74.31$ & $57.06$ & $70.31$ & $75.15$ & $56.89$ & $32.13$ & $46.67$ & $\first{5.32}$  &  $71.11$  &  $74.38$\\\hdashline
\textbf{AdaFace ($m=0.4$)}          &  \scalebox{0.86}{WebFace4M~\cite{zhu2021webface260m}}      &  $\first{70.42}$ & $\first{75.29}$ & $\first{58.27}$ & $\first{70.93}$ & $\first{76.11}$ & $\first{58.02}$ & $\first{35.05}$ & $\first{48.22}$ & $4.96$  &  $\first{72.02}$  &  $\first{74.52}$\\\hline
\end{tabular}
}
\small
\\(b) A performance comparison of recent methods on low quality datasets. 
\scriptsize
\vspace{-2mm}
\caption{Comparison on benchmark datasets, with the ResNet100 backbone.
For high quality and mixed quality datasets, $1$:$1$ verification accuracy and TAR@FAR=$0.01\%$ are reported respectively.
For IJB-S, 
 open-set TPIR@FPIR=$1\%$ and closed-set rank retrieval (Rank-$1$ and Rank-$5$) are reported.
Rank retrieval is also used for TinyFace.
[KEYS: \firstkey{Best}, \secondkey{Second best}, *=our evaluation of the released model]  \vspace{-3mm}}
\label{SOTA_table}
\end{table*}

To compare with SoTA methods, we evaluate ResNet100 trained with AdaFace loss on $9$ datasets as listed in Sec.~\ref{implementation}. 
For the high quality datasets, Tab.~\ref{SOTA_table} (a) shows that AdaFace performs on par with competitive methods such as BroadFace~\cite{kim2020broadface}, SCF-ArcFace~\cite{li2021spherical} and VPL-ArcFace~\cite{deng2021variational}.
This strong performance in high quality datasets is due to the hard sample emphasis on high quality cases during training. 
Note that some performances in high quality datasets are saturated, making the gain less pronounced.
Thus, choosing one model over the others is somewhat difficult based solely on the numbers. 
Unlike SCF-ArcFace, our method does not use additional learnable layers, nor requires $2$-stage training. 
It is a revamp of the loss function, which makes it easier to apply our method to new tasks or backbones. 

For mixed quality datasets, Tab.~\ref{SOTA_table} (a) clearly shows the improvement of AdaFace.
On IJB-B and IJB-C, AdaFace reduces the errors of the second best relatively by $11\%$ and $9\%$ respectively. 
This shows the efficacy of using feature norms as an image quality proxy to treat samples differently. 

For low quality datasets, 
Tab.~\ref{SOTA_table} (b) shows that AdaFace substantially outperforms all baselines.
Compared to the second best, our averaged performance gain over $4$ Rank-$1$ metrics is $3.5\%$, and over $3$ TPIR@=FPIR=1\% metrics is $2.4\%$.
These results show that AdaFace is effective in learning a good representation for the low quality settings as it prevents the model from fitting on unidentifiable images.

We further train on a refined dataset, MS1MV3~\cite{deng2019lightweight} for a fair comparison with a recent work VPL-ArcFace~\cite{deng2021variational}. 
The performance using MS1MV3 is higher than MS1MV2 due to less noise in MS1MV3. 
We also train on newly released WebFace4M~\cite{zhu2021webface260m} dataset.
While one method might shine on one type of data, it is remarkable to see that collectively Adaface achieves SOTA performance on test data with a wide range of image quality, and on various training sets.

\Section{Conclusion}
In this work, we address the problem arising from unidentifiable face images in the training dataset. Data collection processes or data augmentations introduce these images in the training data. Motivated by the difference in recognizability based on image quality, we tackle the problem by 1) using a feature norm as a proxy for the image quality and 2) changing the margin function adaptively based on the feature norm to control the gradient scale assigned to different quality of images. We evaluate the efficacy of the proposed adaptive loss on various qualities of datasets and achieve SoTA for mixed and low quality face datasets.

\Paragraph{Limitations}
This work addresses the existence of unidentifiable images in the training data. 
However, a noisy label is also one of the prominent characteristics of large-scale facial training datasets. 
Our loss function does not give special treatment to mislabeled samples. 
Since our adaptive loss assigns large importance to difficult samples of high quality, high quality mislabeled images can be wrongly emphasized. 
We believe future works may adaptively handle both unidentifiability and label noise at the same time.

\vspace{-0.5mm}
\Paragraph{Potential Societal Impacts}
We believe that the Computer Vision community as a whole should strive to minimize the negative societal impact. 
Our experiments use the training dataset MS1MV*, which is a by-product of MS-Celeb\cite{celeba}, a dataset withdrawn by its creator. 
Our usage of MS1MV* is necessary to compare our result with SoTA methods on a fair basis. 
However, we believe the community should move to new datasets, so we include results on newly released WebFace4M~\cite{zhu2021webface260m}, to facilitate future research. 
In the scientific community, collecting human data requires IRB approval to ensure informed consent.
While IRB status is typically not provided by dataset creators, we assume that most FR datasets (with the exceptions of IJB-S) do not have IRB, due to the nature of collection procedures. 
One direction of the FR community is to collect large datasets with informed consent, fostering R\&D without societal concerns.

\newpage

{\small
\bibliographystyle{ieee_fullname}
\bibliography{egbib}

\begin{thebibliography}{10}\itemsep=-1pt

\bibitem{insightface}
{{I}nsight{F}ace}.
\newblock \url{https://github.com/deepinsight/insightface.git}.
\newblock Accessed: 2021-9-1.

\bibitem{InsightFace_Pytorch}
{{I}nsightFace{P}ytorch}.
\newblock \url{https://github.com/TreB1eN/InsightFace_Pytorch.git}.
\newblock Accessed: 2021-9-1.

\bibitem{TFace}
{{TFace}}.
\newblock \url{https://github.com/Tencent/TFace.git}.
\newblock Accessed: 2021-10-3.

\bibitem{bengio2009curriculum}
Yoshua Bengio, J{\'e}r{\^o}me Louradour, Ronan Collobert, and Jason Weston.
\newblock Curriculum learning.
\newblock In {\em Proceedings of the 26th Annual International Conference on
  Machine Learning}, pages 41--48, 2009.

\bibitem{chang2020data}
Jie Chang, Zhonghao Lan, Changmao Cheng, and Yichen Wei.
\newblock Data uncertainty learning in face recognition.
\newblock In {\em Proceedings of the IEEE/CVF Conference on Computer Vision and
  Pattern Recognition}, pages 5710--5719, 2020.

\bibitem{tinyface}
Zhiyi Cheng, Xiatian Zhu, and Shaogang Gong.
\newblock Low-resolution face recognition.
\newblock In {\em Asian Conference on Computer Vision}, pages 605--621, 2018.

\bibitem{deng2019arcface}
Jiankang Deng, Jia Guo, Niannan Xue, and Stefanos Zafeiriou.
\newblock {ArcFace}: Additive angular margin loss for deep face recognition.
\newblock In {\em Proceedings of the IEEE/CVF Conference on Computer Vision and
  Pattern Recognition}, pages 4690--4699, 2019.

\bibitem{deng2021variational}
Jiankang Deng, Jia Guo, Jing Yang, Alexandros Lattas, and Stefanos Zafeiriou.
\newblock Variational prototype learning for deep face recognition.
\newblock In {\em Proceedings of the IEEE/CVF Conference on Computer Vision and
  Pattern Recognition}, pages 11906--11915, 2021.

\bibitem{deng2019lightweight}
Jiankang Deng, Jia Guo, Debing Zhang, Yafeng Deng, Xiangju Lu, and Song Shi.
\newblock Lightweight face recognition challenge.
\newblock In {\em Proceedings of the IEEE/CVF International Conference on
  Computer Vision Workshops}, pages 0--0, 2019.

\bibitem{feng2018joint}
Yao Feng, Fan Wu, Xiaohu Shao, Yanfeng Wang, and Xi Zhou.
\newblock Joint 3d face reconstruction and dense alignment with position map
  regression network.
\newblock In {\em European Conference on Computer Vision}, pages 534--551,
  2018.

\bibitem{msceleb}
Yandong Guo, Lei Zhang, Yuxiao Hu, Xiaodong He, and Jianfeng Gao.
\newblock {MS-Celeb-1M}: A dataset and benchmark for large-scale face
  recognition.
\newblock In {\em European Conference on Computer Vision}, pages 87--102, 2016.

\bibitem{he2016deep}
Kaiming He, Xiangyu Zhang, Shaoqing Ren, and Jian Sun.
\newblock Deep residual learning for image recognition.
\newblock In {\em Proceedings of the IEEE Conference on Computer Vision and
  Pattern Recognition}, pages 770--778, 2016.

\bibitem{he2019bag}
Tong He, Zhi Zhang, Hang Zhang, Zhongyue Zhang, Junyuan Xie, and Mu Li.
\newblock Bag of tricks for image classification with convolutional neural
  networks.
\newblock In {\em Proceedings of the IEEE/CVF Conference on Computer Vision and
  Pattern Recognition}, pages 558--567, 2019.

\bibitem{lfw}
Gary~B Huang, Marwan Mattar, Tamara Berg, and Eric Learned-Miller.
\newblock Labeled {Faces} in the {Wild}: A database forstudying face
  recognition in unconstrained environments.
\newblock In {\em Workshop on Faces in'Real-Life'Images: Detection, Alignment,
  and Recognition}, 2008.

\bibitem{huang2020improving}
Yuge Huang, Pengcheng Shen, Ying Tai, Shaoxin Li, Xiaoming Liu, Jilin Li,
  Feiyue Huang, and Rongrong Ji.
\newblock Improving face recognition from hard samples via distribution
  distillation loss.
\newblock In {\em European Conference on Computer Vision}, pages 138--154,
  2020.

\bibitem{huang2020curricularface}
Yuge Huang, Yuhan Wang, Ying Tai, Xiaoming Liu, Pengcheng Shen, Shaoxin Li,
  Jilin Li, and Feiyue Huang.
\newblock {CurricularFace}: adaptive curriculum learning loss for deep face
  recognition.
\newblock In {\em Proceedings of the IEEE/CVF Conference on Computer Vision and
  Pattern Recognition}, pages 5901--5910, 2020.

\bibitem{ijbs}
Nathan~D Kalka, Brianna Maze, James~A Duncan, Kevin O’Connor, Stephen
  Elliott, Kaleb Hebert, Julia Bryan, and Anil~K Jain.
\newblock {IJB--S}: {IARPA} {Janus} {Surveillance} {Video} {Benchmark}.
\newblock In {\em 2018 IEEE 9th International Conference on Biometrics Theory,
  Applications and Systems (BTAS)}, pages 1--9, 2018.

\bibitem{kang2019attentional}
Bong-Nam Kang, Yonghyun Kim, Bongjin Jun, and Daijin Kim.
\newblock Attentional feature-pair relation networks for accurate face
  recognition.
\newblock In {\em Proceedings of the IEEE/CVF International Conference on
  Computer Vision}, pages 5472--5481, 2019.

\bibitem{kim2021quality}
Insoo Kim, Seungju Han, Ji-won Baek, Seong-Jin Park, Jae-Joon Han, and Jinwoo
  Shin.
\newblock Quality-agnostic image recognition via invertible decoder.
\newblock In {\em Proceedings of the IEEE/CVF Conference on Computer Vision and
  Pattern Recognition}, pages 12257--12266, 2021.

\bibitem{kim2020broadface}
Yonghyun Kim, Wonpyo Park, and Jongju Shin.
\newblock {BroadFace}: Looking at tens of thousands of people at once for face
  recognition.
\newblock In {\em European Conference on Computer Vision}, pages 536--552,
  2020.

\bibitem{li2021spherical}
Shen Li, Jianqing Xu, Xiaqing Xu, Pengcheng Shen, Shaoxin Li, and Bryan Hooi.
\newblock Spherical confidence learning for face recognition.
\newblock In {\em Proceedings of the IEEE/CVF Conference on Computer Vision and
  Pattern Recognition}, pages 15629--15637, 2021.

\bibitem{lin2017focal}
Tsung-Yi Lin, Priya Goyal, Ross Girshick, Kaiming He, and Piotr Doll{\'a}r.
\newblock Focal loss for dense object detection.
\newblock In {\em Proceedings of the IEEE International Conference on Computer
  Vision}, pages 2980--2988, 2017.

\bibitem{liu2021dam}
Jiaheng Liu, Yudong Wu, Yichao Wu, Chuming Li, Xiaolin Hu, Ding Liang, and
  Mengyu Wang.
\newblock {DAM}: Discrepancy alignment metric for face recognition.
\newblock In {\em Proceedings of the IEEE/CVF International Conference on
  Computer Vision}, pages 3814--3823, 2021.

\bibitem{liu2017sphereface}
Weiyang Liu, Yandong Wen, Zhiding Yu, Ming Li, Bhiksha Raj, and Le Song.
\newblock {SphereFace}: Deep hypersphere embedding for face recognition.
\newblock In {\em Proceedings of the IEEE Conference on Computer Vision and
  Pattern Recognition}, pages 212--220, 2017.

\bibitem{celeba}
Ziwei Liu, Ping Luo, Xiaogang Wang, and Xiaoou Tang.
\newblock Deep learning face attributes in the wild.
\newblock In {\em Proceedings of the IEEE International Conference on Computer
  Vision}, pages 3730--3738, 2015.

\bibitem{ijbc}
Brianna Maze, Jocelyn Adams, James~A Duncan, Nathan Kalka, Tim Miller, Charles
  Otto, Anil~K Jain, W~Tyler Niggel, Janet Anderson, Jordan Cheney, and Patrick
  Grother.
\newblock {IARPA} {Janus} {Benchmark}-{C}: Face dataset and protocol.
\newblock In {\em 2018 International Conference on Biometrics (ICB)}, pages
  158--165, 2018.

\bibitem{meng2021magface}
Qiang Meng, Shichao Zhao, Zhida Huang, and Feng Zhou.
\newblock {MagFace}: A universal representation for face recognition and
  quality assessment.
\newblock In {\em Proceedings of the IEEE/CVF Conference on Computer Vision and
  Pattern Recognition}, pages 14225--14234, 2021.

\bibitem{brisque}
Anish Mittal, Anush~Krishna Moorthy, and Alan~Conrad Bovik.
\newblock No-reference image quality assessment in the spatial domain.
\newblock {\em IEEE Transactions on Image Processing}, 21(12):4695--4708, 2012.

\bibitem{agedb}
Stylianos Moschoglou, Athanasios Papaioannou, Christos Sagonas, Jiankang Deng,
  Irene Kotsia, and Stefanos Zafeiriou.
\newblock {AGEDB}: the first manually collected, in-the-wild age database.
\newblock In {\em Proceedings of the IEEE Conference on Computer Vision and
  Pattern Recognition Workshops}, pages 51--59, 2017.

\bibitem{improving-face-recognition-with-a-quality-based-probabilistic-framework}
Necmiye Ozay, Yan Tong, Frederick~W. Wheeler, and Xiaoming Liu.
\newblock Improving face recognition with a quality-based probabilistic
  framework.
\newblock In {\em Proceeding of IEEE Conference on Computer Vision and Pattern
  Recognition Workshops}, pages 134--141, 2009.

\bibitem{cfpfp}
Soumyadip Sengupta, Jun-Cheng Chen, Carlos Castillo, Vishal~M Patel, Rama
  Chellappa, and David~W Jacobs.
\newblock Frontal to profile face verification in the wild.
\newblock In {\em 2016 IEEE Winter Conference on Applications of Computer
  Vision (WACV)}, pages 1--9, 2016.

\bibitem{sheikh2006image}
Hamid~R Sheikh and Alan~C Bovik.
\newblock Image information and visual quality.
\newblock {\em IEEE Transactions on Image Processing}, 15(2):430--444, 2006.

\bibitem{shi2019probabilistic}
Yichun Shi and Anil~K Jain.
\newblock Probabilistic face embeddings.
\newblock In {\em Proceedings of the IEEE/CVF International Conference on
  Computer Vision}, pages 6902--6911, 2019.

\bibitem{shi2020towards}
Yichun Shi, Xiang Yu, Kihyuk Sohn, Manmohan Chandraker, and Anil~K Jain.
\newblock Towards universal representation learning for deep face recognition.
\newblock In {\em Proceedings of the IEEE/CVF Conference on Computer Vision and
  Pattern Recognition}, pages 6817--6826, 2020.

\bibitem{shrivastava2016training}
Abhinav Shrivastava, Abhinav Gupta, and Ross Girshick.
\newblock Training region-based object detectors with online hard example
  mining.
\newblock In {\em Proceedings of the IEEE Conference on Computer Vision and
  Pattern Recognition}, pages 761--769, 2016.

\bibitem{serfiq}
Philipp Terh{\"o}rst, Jan~Niklas Kolf, Naser Damer, Florian Kirchbuchner, and
  Arjan Kuijper.
\newblock Ser-fiq: unsupervised estimation of face image quality based on
  stochastic embedding robustness. in 2020 ieee.
\newblock In {\em CVF Conference on Computer Vision and Pattern Recognition,
  CVPR}, pages 13--19, 2020.

\bibitem{disentangled-representation-learning-gan-for-pose-invariant-face-recognition}
Luan Tran, Xi Yin, and Xiaoming Liu.
\newblock Disentangled representation learning {GAN} for pose-invariant face
  recognition.
\newblock In {\em Proceeding of IEEE Computer Vision and Pattern Recognition},
  pages 1415--1424, 2017.

\bibitem{wang2017normface}
Feng Wang, Xiang Xiang, Jian Cheng, and Alan~Loddon Yuille.
\newblock {NormFace}: L2 hypersphere embedding for face verification.
\newblock In {\em Proceedings of the 25th ACM International Conference on
  Multimedia}, pages 1041--1049, 2017.

\bibitem{wang2018cosface}
Hao Wang, Yitong Wang, Zheng Zhou, Xing Ji, Dihong Gong, Jingchao Zhou, Zhifeng
  Li, and Wei Liu.
\newblock {CosFace}: Large margin cosine loss for deep face recognition.
\newblock In {\em Proceedings of the IEEE Conference on Computer Vision and
  Pattern Recognition}, pages 5265--5274, 2018.

\bibitem{wang2020mis}
Xiaobo Wang, Shifeng Zhang, Shuo Wang, Tianyu Fu, Hailin Shi, and Tao Mei.
\newblock {Mis-classified} vector guided softmax loss for face recognition.
\newblock In {\em Proceedings of the AAAI Conference on Artificial
  Intelligence}, volume~34, pages 12241--12248, 2020.

\bibitem{ijbb}
Cameron Whitelam, Emma Taborsky, Austin Blanton, Brianna Maze, Jocelyn Adams,
  Tim Miller, Nathan Kalka, Anil~K Jain, James~A Duncan, Kristen Allen, et~al.
\newblock {IARPA} {Janus} {Benchmark}-{B} face dataset.
\newblock In {\em Proceedings of the IEEE Conference on Computer Vision and
  Pattern Recognition Workshops}, pages 90--98, 2017.

\bibitem{yin2020fan}
Xi Yin, Ying Tai, Yuge Huang, and Xiaoming Liu.
\newblock {FAN}: Feature adaptation network for surveillance face recognition
  and normalization.
\newblock In {\em Proceedings of the Asian Conference on Computer Vision},
  pages 301--319, 2020.

\bibitem{zhai2020perceptual}
Guangtao Zhai and Xiongkuo Min.
\newblock Perceptual image quality assessment: a survey.
\newblock {\em Science China Information Sciences}, 63(11):211301, 2020.

\bibitem{zhang2016joint}
Kaipeng Zhang, Zhanpeng Zhang, Zhifeng Li, and Yu Qiao.
\newblock Joint face detection and alignment using multitask cascaded
  convolutional networks.
\newblock {\em IEEE Signal Processing Letters}, 23(10):1499--1503, 2016.

\bibitem{zhang2019adacos}
Xiao Zhang, Rui Zhao, Yu Qiao, Xiaogang Wang, and Hongsheng Li.
\newblock Adacos: Adaptively scaling cosine logits for effectively learning
  deep face representations.
\newblock In {\em Proceedings of the IEEE/CVF Conference on Computer Vision and
  Pattern Recognition}, pages 10823--10832, 2019.

\bibitem{zhang2019p2sgrad}
Xiao Zhang, Rui Zhao, Junjie Yan, Mengya Gao, Yu Qiao, Xiaogang Wang, and
  Hongsheng Li.
\newblock {P2sGrad}: Refined gradients for optimizing deep face models.
\newblock In {\em Proceedings of the IEEE/CVF Conference on Computer Vision and
  Pattern Recognition}, pages 9906--9914, 2019.

\bibitem{zheng2020automatic}
Jingxiao Zheng, Rajeev Ranjan, Ching-Hui Chen, Jun-Cheng Chen, Carlos~D
  Castillo, and Rama Chellappa.
\newblock An automatic system for unconstrained video-based face recognition.
\newblock {\em IEEE Transactions on Biometrics, Behavior, and Identity
  Science}, 2(3):194--209, 2020.

\bibitem{zheng2019uncertainty}
Jingxiao Zheng, Ruichi Yu, Jun-Cheng Chen, Boyu Lu, Carlos~D Castillo, and Rama
  Chellappa.
\newblock Uncertainty modeling of contextual-connections between tracklets for
  unconstrained video-based face recognition.
\newblock In {\em Proceedings of the IEEE/CVF International Conference on
  Computer Vision}, pages 703--712, 2019.

\bibitem{cplfw}
Tianyue Zheng and Weihong Deng.
\newblock Cross-{Pose} {LFW}: A database for studying cross-pose face
  recognition in unconstrained environments.
\newblock {\em Beijing University of Posts and Telecommunications, Tech. Rep},
  5:7, 2018.

\bibitem{calfw}
Tianyue Zheng, Weihong Deng, and Jiani Hu.
\newblock Cross-{Age} {LFW:} {A} database for studying cross-age face
  recognition in unconstrained environments.
\newblock {\em CoRR}, abs/1708.08197, 2017.

\bibitem{zhou2003probabilistic}
Shaohua Zhou, Volker Krueger, and Rama Chellappa.
\newblock Probabilistic recognition of human faces from video.
\newblock {\em CVIU}, 91(1-2):214--245, 2003.

\bibitem{zhu2021webface260m}
Zheng Zhu, Guan Huang, Jiankang Deng, Yun Ye, Junjie Huang, Xinze Chen, Jiagang
  Zhu, Tian Yang, Jiwen Lu, Dalong Du, and Jie Zhou.
\newblock {WebFace260M}: A benchmark unveiling the power of million-scale deep
  face recognition.
\newblock In {\em Proceedings of the IEEE/CVF Conference on Computer Vision and
  Pattern Recognition}, pages 10492--10502, 2021.

\end{thebibliography}
}


\onecolumn
\setcounter{equation}{0}
\setcounter{figure}{0}
\setcounter{table}{0}
\setcounter{page}{1}
\setcounter{section}{0}

\begin{center}
\textbf{\Large AdaFace: Quality Adaptive Margin for Face Recognition}\\
\vspace{2mm}
\textbf{\large Supplementary Material}\\
\end{center}

\renewcommand\thesection{\Alph{section}}


\renewcommand\thesection{\Alph{section}}

\section{Gradient Scaling Term}
In Sec.~3.1 of the main paper, the gradient scaling term (GST), $g$ is introduced. Specifically, it is derived from the gradient equation for the margin-based softmax loss and defined as 
\begin{equation}
    g := \left( P_j^{(i)} - \mathbbm{1}(y_i = j) \right) \frac{\partial f(\cos \theta_j) }{ \partial \cos \theta_j},
    \label{gradient_scale_def2}
\end{equation}
where 
\begin{equation}
    P_j^{(i)} = \frac{\exp(f(\cos \theta_{y_i}) )}{\exp(f(\cos \theta_{y_i})) + \sum_{j\neq y_i}^n \exp(s\cos\theta_{j})}.
    \label{def_p}
\end{equation}

This scalar term, $g$ affects the magnitude of the gradient during backpropagation from the margin-based softmax loss. The form of $g$ depends on the form of the margin function $f(\cos\theta_j)$. In Tab.~\ref{table:margin_summary}, we summarize the margin function $f(\cos\theta_j)$ and the corresponding GST when $j=y_{i}$, the ground truth index.

\begin{table}[h]
\centering
\small
\setlength{\tabcolsep}{3pt}
\begin{tabular}{|l|c|c|c|}
\hline
Methods    & $f(\cos \theta_{j})$, $j\neq y_i$  & $f(\cos \theta_{j})$, $j= y_i$  & $g$ when $j=y_{i}$ \\ \hline\hline
Softmax   &   $s \cdot \cos \theta_{y_i}$  &  $s \cdot \cos \theta_{y_i}$     &  $\left(P_{y_i}^{(i)}-1\right) s$             \\ \hline \hline
Additive Margin (CosFace~\cite{wang2018cosface})    &   $s \cdot \cos \theta_{y_i}$  &  $s (\cos \theta_{y_i} - m)$      &    $\left(P_{y_i}^{(i)}-1\right) s $          \\ \hline \hline
Angular Margin (ArcFace~\cite{deng2019arcface})    &   $s \cdot \cos \theta_{y_i}$  &  $s \cdot \cos (\theta_{y_i} + m)$      &        $\left(P_{y_i}^{(i)}-1\right) s \left( \cos(m) + \frac{\cos \theta_{y_i} \sin(m)}{\sqrt{1 - \cos^2 \theta_{y_i}}} \right) $    \\ 
\hline \hline
\multicolumn{1}{|l|}{\multirow{2}{*}{Adaptive Angular Margin}}   &  $s \cdot \cos \theta_{y_i}$  &  $s \cdot \cos (\theta_{y_i} + m(\Vert z_i\Vert))$      &        $\left(P_{y_i}^{(i)}-1\right) s \left( \cos(m(\Vert z_i\Vert)) + \frac{\cos \theta_{y_i} \sin(m(\Vert z_i\Vert))}{\sqrt{1 - \cos^2 \theta_{y_i}}} \right) $        \\ \cline{2-4} 
& \multicolumn{3}{c|}{\multirow{1}{*}{$m(\Vert z_i\Vert)= \text{a monotonically inc. function of } \Vert z_i \Vert$. In this table, $g$ is derived with $ \Vert z_i \Vert$ as a constant.  }} \\ \hline \hline
\multicolumn{1}{|l|}{\multirow{2}{*}{CurricularFace~\cite{huang2020curricularface}}} &   $N(t, \cos \theta_j)$  &   $s \cdot \cos (\theta_{y_i} + m)$    &        $\left(P_{y_i}^{(i)}-1\right) s \left( \cos(m) + \frac{\cos \theta_{y_i} \sin(m)}{\sqrt{1 - \cos^2 \theta_{y_i}}} \right) $    \\ 
\cline{2-4}
& \multicolumn{3}{c|}{\multirow{1}{*}{$N(t, \cos \theta_j) = \cos (\theta_{j})(t+\cos\theta_j) \text{\quad if \quad } s \cos (\theta_{y_i}+m)  < \cos\theta_j \text{\quad else \quad } \cos (\theta_{j})$}}  \\ \hline \hline
\multicolumn{1}{|l|}{\multirow{2}{*}{AdaFace (ours)}}    &  $s \cdot \cos \theta_{y_i}$  &  $s \cdot \cos (\theta_{y_i} + g_{\text{angle}} ) - g_{\text{add}}$      &        $\left(P_{y_i}^{(i)}-1\right) s \left( \cos(g_{\text{angle}}) + \frac{\cos \theta_{y_i} \sin(g_{\text{angle}})}{\sqrt{1 - \cos^2 \theta_{y_i}}} \right) $        \\ \cline{2-4}
& \multicolumn{2}{c|}{\multirow{1}{*}{$g_{\text{angle}} = -m \cdot \widehat{\Vert \bm{z}_i \Vert},\quad
    g_{\text{add}} = m \cdot \widehat{\Vert \bm{z}_i \Vert} + m$}}  & $\widehat{\Vert \bm{z}_i \Vert} = \clip{\frac{\Vert \bm{z}_i \Vert - \mu_{z}}{\sigma_{z} / h}}^1_{-1}$ \\ \hline
\end{tabular}
\caption{Table of margin functions and their gradient scale terms. The concept of Adaptive Angular Margin is explored in MagFace~\cite{meng2021magface}. However, unlike other works, MagFace is treating $m(\Vert z_i\Vert)$ as a term to optimize (\textit{i.e.} $\Vert z_i \Vert$ is a function of $\cos \theta_{j}$), as oppose to treating it as a constant. In this table, we treat $\Vert z_i\Vert$ as a constant to highlight the effect of the margin. The exact form of $g$ for MagFace will be different. In Fig.~3 of the main paper, Adaptive Angular Margin is visualized using the equation from this table. }
\label{table:margin_summary}
\end{table}

Note that $P_{y_i}$ is also affected by the choice of the margin function $f(\cos\theta_{y_i})$ as in Eqn.~\ref{def_p}. So, $g$ is a function of $m$, except for Softmax, and $g$ is affected by $m$ through $f(\cos\theta_{y_i})$ in $P_{y_i}$. For Angular Margin, $m$ appears in the equation for $g$ directly. We derive $g$ for Angular Margin below. The term $g$ for the Adaptive Angular Margin and CurricularFace~\cite{huang2020curricularface} can be obtained using the $g$ from the Angular Margin. The GST term for AdaFace can be obtained by using $g$ for the Angular Margin and the Additive Margin, and replacing $m$ with adaptive terms $g_{\text{angle}}$ and $g_{\text{add}}$. This is possible because $ \Vert z_i \Vert$ is treated as a constant.

\subsection{Derivation of Angular Margin}
We can rewrite $f(\cos \theta_{y_i})$ as 
\begin{equation}
\begin{split}
f(\cos \theta_{y_i})  & =s \cdot (\cos (\theta_{y_i} + m) ) \\
 & = s \cdot (\cos\theta_{y_i} \cos m - \sin \theta_{y_i} \sin m  )  \\
 &= s \cdot \left( \cos \theta_{y_i} \cos m - \sqrt{1-\cos^2 \theta_{y_i}} \sin m \right),
\end{split}
\end{equation}
by the laws of trignometry. Therefore, 
\begin{equation}
    \frac{\partial f(\cos \theta_{y_i}) }{ \partial \cos \theta_{y_i}}  = s \left( \cos(m) + \frac{\cos \theta_{y_i} \sin(m)}{\sqrt{1 - \cos^2 \theta_{y_i}}} \right). 
\label{angular}
\end{equation}

\subsection{Interpretation of $g$}

\begin{figure}[!h]
\centering
  \includegraphics[width=1.0\linewidth]{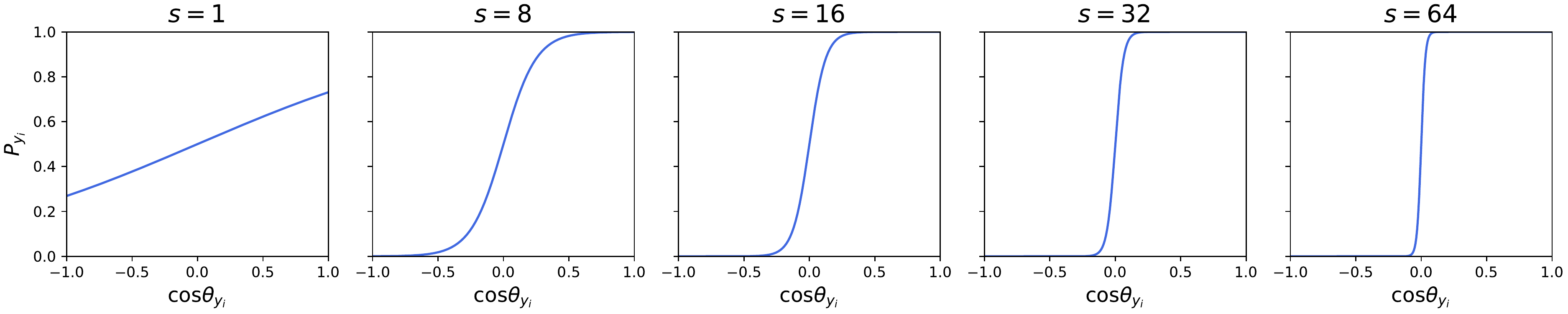}
  \caption{Plot of $P_{y_i}$ for different values of $s$. In this figure, $P_{y_i}$ is calculated with $f(\cos\theta_j)$ from Softamx (\textit{i.e.} $m=0$).}
    \label{prob_with_s}
\end{figure}

For Softmax and Additive Margin, we see that $g=(P_{y_i}^{(i)}-1)s$. Since the softmax operation in $P_{y_i}^{(i)}$ has a tendency to scale the result to be close to either $0$ or $1$, the first term in $g$, $( P_j^{(i)} - 1)$ tends to be close to $1$ or $0$ far away from the decision boundary. In the equation for $P_{y_i}$, there is also $s$ which is a scaling hyper-parameter, and is often set to $s=64$~\cite{wang2018cosface, deng2019arcface, liu2017sphereface, huang2020curricularface}. This high $s$ makes the softmax operation even steeper near the decision boundary. This results in almost equal GST for samples away from the decision boundary, regardless of how far they are from the decision boundary. This is evident in Fig.~\ref{prob_with_s}, where the blue curve is flat except near the decision boundary when $s$ is high. 

\begin{figure}[!h]
\centering
  \includegraphics[width=1.0\linewidth]{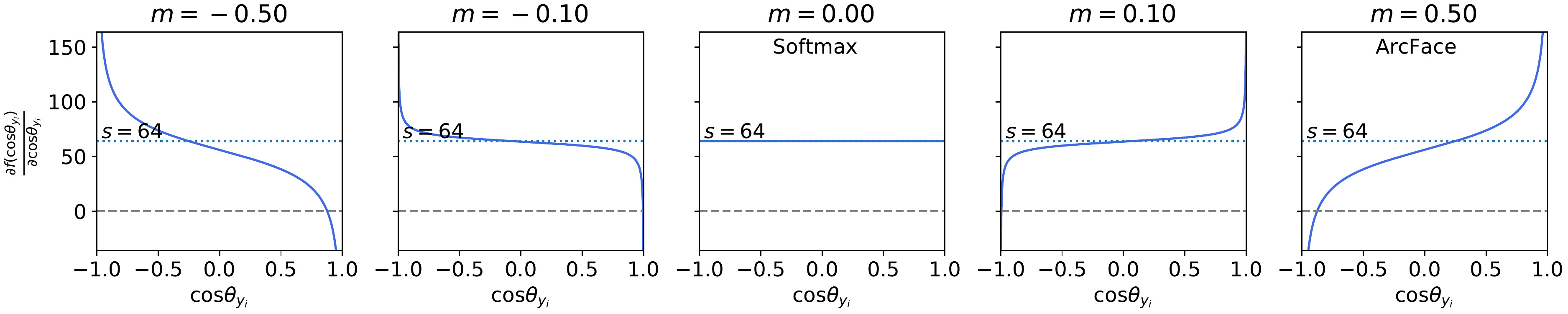}
  \caption{Plot of $\frac{\partial f(\cos \theta_{y_i}) }{ \partial \cos \theta_{y_i}}$ for different value of $m$ when the margin function is Angluar Margin. }
    \label{angular_m}
\end{figure}

For Softmax and Additive Margin,$\frac{\partial f(\cos \theta_{y_i}) }{ \partial \cos \theta_{y_i}} = s$.
This term is different for Angular Margin due to $\frac{\partial f(\cos \theta_{y_i}) }{ \partial \cos \theta_{y_i}}$ being a function of $\cos \theta_{y_i}$. The exact form of $\frac{\partial f(\cos \theta_{y_i}) }{ \partial \cos \theta_{y_i}}$ for Angular Margin is found in  Eqn.~\ref{angular}. As shown in Fig.~\ref{angular_m}, Eqn.~\ref{angular} is monotonically increasing with respect to $\cos\theta_{y_i}$ when $m>0$ and vice versa. Note that $\cos\theta_{y_i}$ is how close the sample is to the ground truth weight vector, and it is closely related to the difficulty of the sample during training. Therefore, this partial derivative term from the angular margin, $\frac{\partial f(\cos \theta_{y_i}) }{ \partial \cos \theta_{y_i}}$, can be viewed as scaling the importance of sample based on the difficulty. 

\newpage
\section{Feature Norm Analysis}
    
    \subsection{Correlation between Norm and BRISQUE during Training}
\begin{figure}[!h]
\centering
  \includegraphics[width=0.84\linewidth]{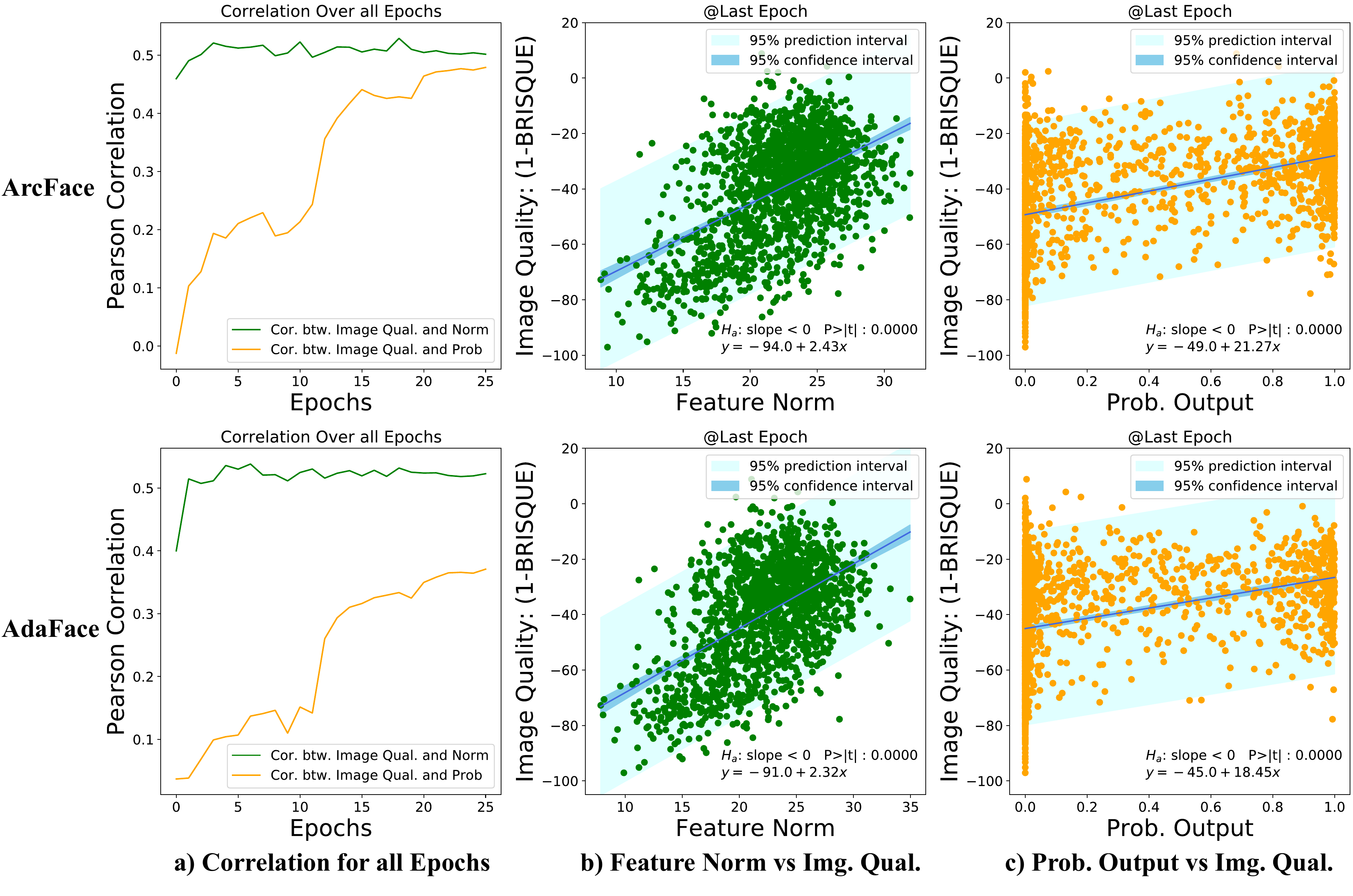}
  \vspace{-2mm}
\caption{Comparison between ArcFace and AdaFace on the correlation between the feature norm and the image quality. We randomly sampled $1,534$ images from the training dataset (MS1MV2~\cite{deng2019arcface}) to show this plot. }
\label{cor_plot}
\end{figure}

In the Sec.~3.2 of the main paper, we introduce the idea of using the feature norm as a proxy of the image quality. We observe that in models trained with a margin-based softmax loss, the feature norm exhibits a trend that is correlated with the image quality. Here, we show for ArcFace and AdaFace, both loss functions exhibit this trend, in Fig.~\ref{cor_plot}. Regardless of the form of the margin function, the correlation between the feature norm and the image quality is quite similar (green plot in 1st and 2nd columns). We leverage this behavior to design the proxy for the image quality.

\begin{figure}
\vspace{-14mm}
    \centering
\begin{minipage}{6in}
  \raisebox{-0.25\height}{\includegraphics[width=0.27\linewidth]{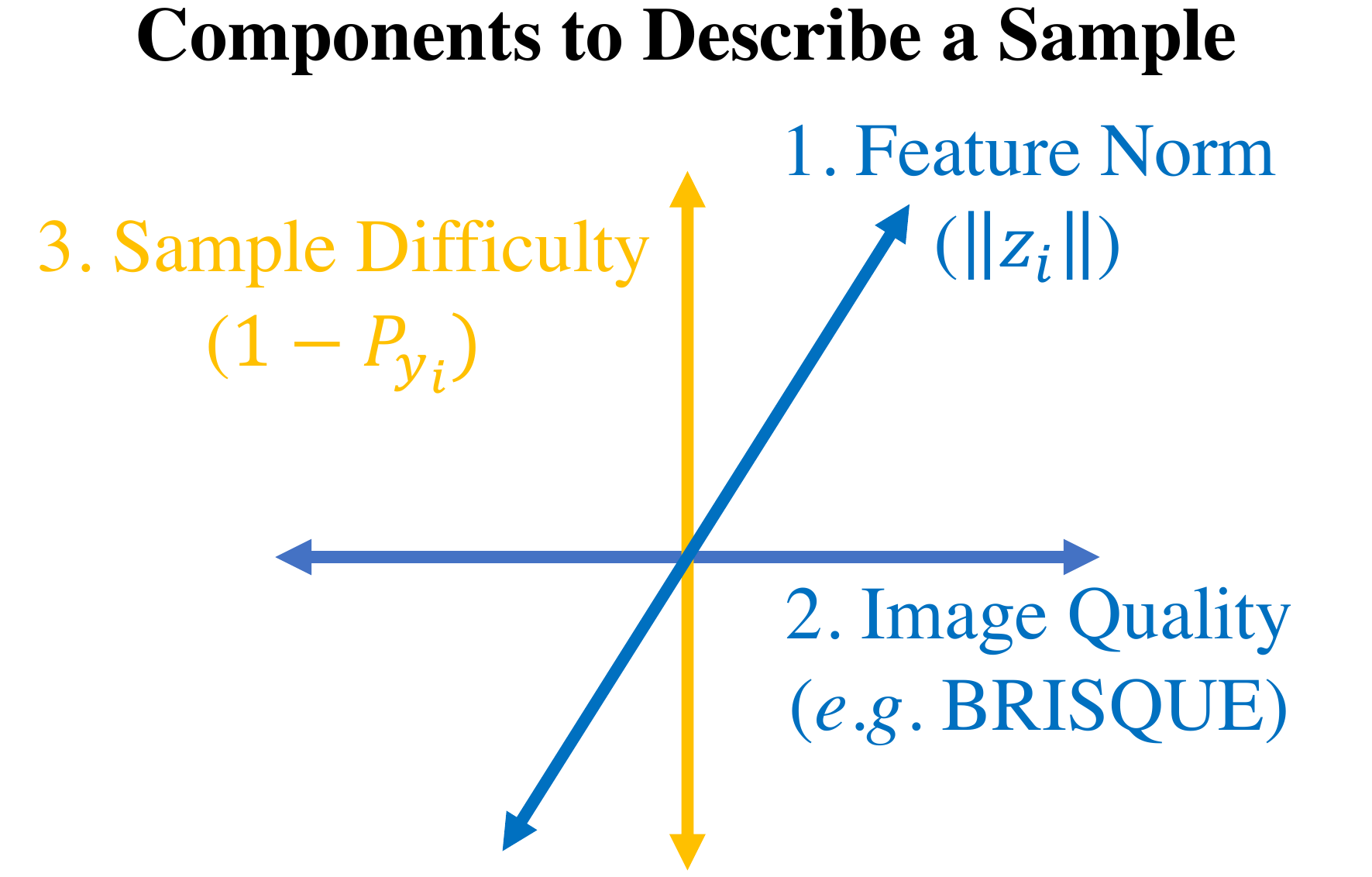}}
  \hspace*{.2in}
  \raisebox{-0.\height}{
  \begin{tabular}[b]{ccc}\hline
      Method & Relationship & Gradient Flow to $\Vert z_i \Vert$ \\ \hline
      MagFace~\cite{meng2021magface} &  Sample Difficulty vs. $\Vert z_i \Vert$ & Yes \\
      AdaFace & Image Qual. vs. $\Vert z_i \Vert$ & No \\ \hline
    \end{tabular}}
\end{minipage}
\vspace{-2mm}
    \caption{An illustration of different components to describe a sample and their usage in previous works.}
    \label{fig:three_components}
\end{figure}

We use three concepts (image quality, feature norm and sample difficulty) to describe a sample, as illustrated in Fig.~\ref{fig:three_components}. We leverage the correlation between the feature norm and the image quality to apply different emphasis to different difficulty of samples. In contrast, MagFace learns a representation that aligns the feature norm with recognizability. The term, \textit{image quality} in MagFace paper~\cite{meng2021magface} refers to the face recognizability, which is closer in meaning to the sample difficulty than the term, image quality, we use in our paper. Please refer to the Fig.~1 (a) and the first contribution claim of the MagFace paper~\cite{meng2021magface}. Also note the difference in gradient flow through the feature norm, $\Vert z_i \Vert$. MagFace relies on learning the feature that has $\Vert z_i \Vert$ aligned with the recognizability of the sample, requiring the gradient to flow through $\Vert z_i \Vert$ during backpropagation. The loss function has the incentive to reduce the margin by reducing $\Vert z_i \Vert$. However, our objective is to adaptively change the loss function, itself, so we treat $\Vert z_i \Vert$ as a constant.
Finally, from Tab.~3 of our main paper, AdaFace substantially outperforms MagFace, {\it e.g.}~reducing the errors of MagFace on IJB-B and IJB-C relatively by $21\%$ and $23\%$ respectively.

\newpage
\subsection{Training Sample Visualization}

\begin{figure}[h]
\centering
  \includegraphics[width=0.85\linewidth]{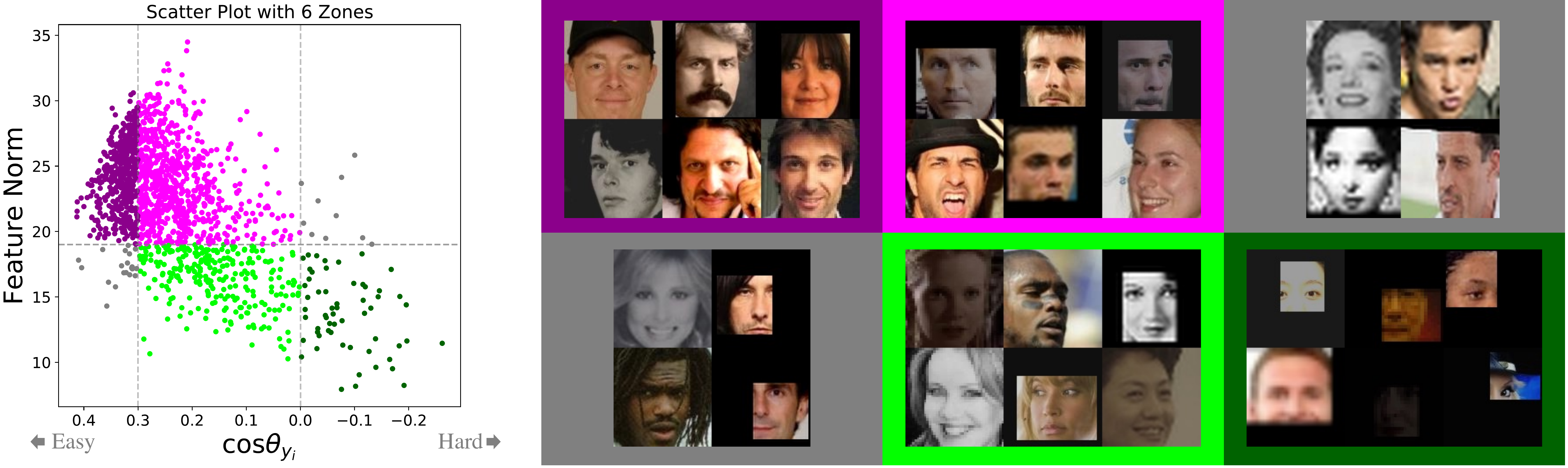}
  \caption{Actual training data examples corresponding to $6$ zones. A pretrained AdaFace model is used as a feature extractor.}
  \label{6zone}
\end{figure}

We show some visualization of the actual training images. From the randomly sampled $1,534$ images from the training dataset (MS1MV2~\cite{deng2019arcface}), we divide the samples into $6$ different zones. We plot the samples by $\cos\theta_{y_i}$ (decreasing) as the x-axis and the feature norm $\Vert z_i\Vert$ as y-axis in Fig.~\ref{6zone}. We divide the plot into $6$ zones and sample a few images from each group. Clearly, there are not many samples in the zones highlighted by the gray area (top right and bottom left). This indicates that the sample difficulty distribution is different for each level of feature norm. 
Furthermore, the samples in the dark green area are mostly unrecognizable images. AdaFace de-emphasizes these samples. 
Also, the samples in the bright pink area are more difficult samples than the dark pink area. 
AdaFace puts more emphasis on the harder samples when the feature norm is high.
We would like to reminde the readers thatthis figure may serve as {\it an empirical validation} of the two-dimensional face image categorization we made in Fig.~1 of the main paper.  

\subsection{Training Samples' Gradient Scaling Term for AdaFace}

\begin{figure}[h]
\centering
  \includegraphics[width=0.9\linewidth]{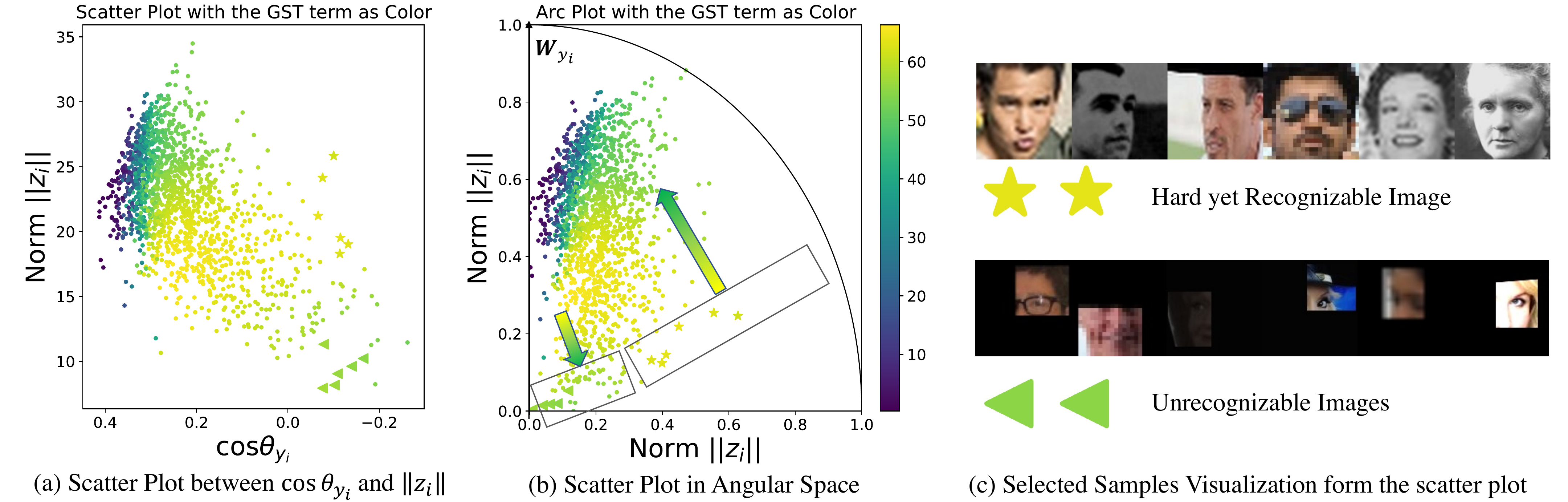}
  \caption{(a) Scatter plot of samples from Fig.~\ref{6zone} with the color as the GST term. (b): Scatter plot of the same $1,534$ points in angular space. For each feature, the angle from $W_{y_i}$ is calculated from $\cos\theta_{y_i}$ and the distance from the origin is calculated from $\Vert z_i \Vert$. Both terms are normalized for visualization. (c): Sample image visualization from the low norm and high norm regions of similar $\cos\theta_{y_i}$.}
  \label{gst_scatter}
\end{figure}

In Fig.~\ref{gst_scatter} (a), we plot the actual GST term for AdaFace. We use the same $1,534$ images from the training dataset (MS1MV2~\cite{deng2019arcface}) as in Fig.~\ref{6zone}.
The color of points indicates the magnitude of the GST term. The purple points on the left side of the scatter plot are samples past the decision boundary. Therefore the magnitude of GST term is low. The effective difference in GST term for samples outside the decision boundary can be seen by the color change from green to yellow. Note that AdaFace de-emphasizes samples of low feature norm and high difficulty. This is shown in the lower right region of the plot. 
In Fig.~\ref{gst_scatter} (b), we warp the plot into the angular space to make a correspondence with the Fig.~3 of the main paper, where we illustrate the GST term for AdaFace. We illustrate how actual training samples are distributed in this angular space. 
In Fig.~\ref{gst_scatter} (b) and (c), we visualize two groups of images where one is from the low feature norm area (triangle) and the other is from the high feature norm area (star). AdaFace exploits images that are hard yet recognizable, as indicated by the yellow star regions, and lowers the learning signal from the unrecognizable images, as indicated by the green triangle regions.  

\subsection{Train Samples' Gradient Scaling Term Comparison with ArcFace}
\begin{figure}[h]
\centering
  \includegraphics[width=1.0\linewidth]{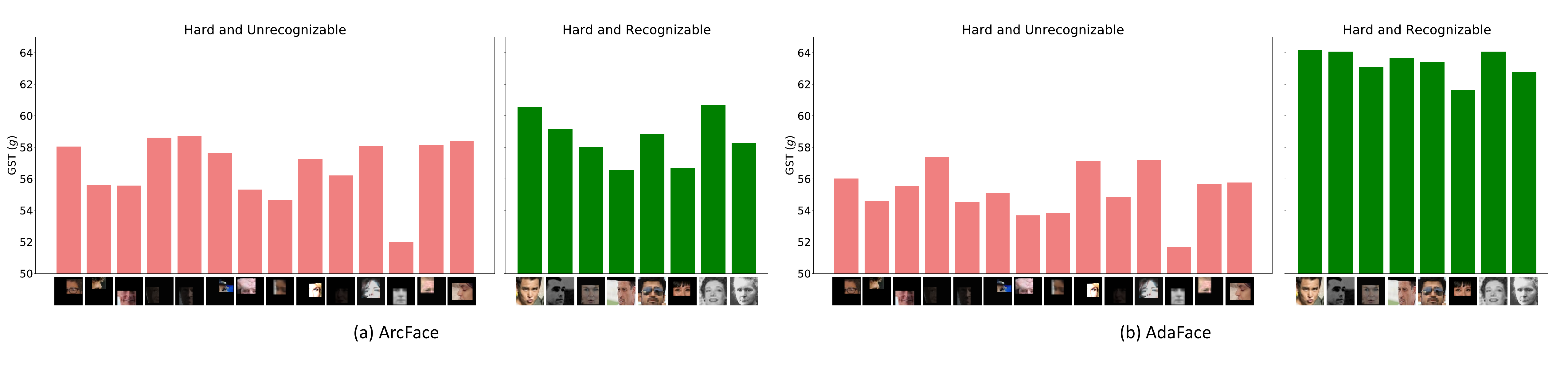}
  \caption{Comparison of the magnitude of GST term between ArcFace and AdaFace.}
  \label{arcface_adaface_compare}
\end{figure}

In Fig.~\ref{arcface_adaface_compare}, we compare the GST term placed on training samples. We have two groups of images. One group is comprised of unrecognizable images, shown under the red bar. Another group is comprised of hard yet recognizable images, shown under the green bar. Each bar corresponds to one training sample, and the height of the bar indicates the magnitude of the gradient scaling term (GST). For ArcFace shown on the left, the same level of GST is placed on all samples. However, in AdaFace, unrecognizable samples are less emphasized relative to the recognizable samples.

\section{Visualization of Success and Failed Test Images }
We show samples from IJB-C~\cite{ijbc} dataset to show which samples are correctly classified in AdaFace, compared to ArcFace~\cite{deng2019arcface}. In each pair of probe and gallery images, we write the rank and the similarity score for both ArcFace and AdaFace. Rank$=1$ is the correct match and a high similarity score is desired. Note that the majority of the cases where AdaFace successfully matches the hard samples for ArcFace are comprised of low quality samples. This shows that indeed AdaFace works well on low quality images.

\begin{figure}[!ht]
    \centering
    \begin{minipage}{0.45\textwidth}
        \centering
        \includegraphics[width=7.6cm]{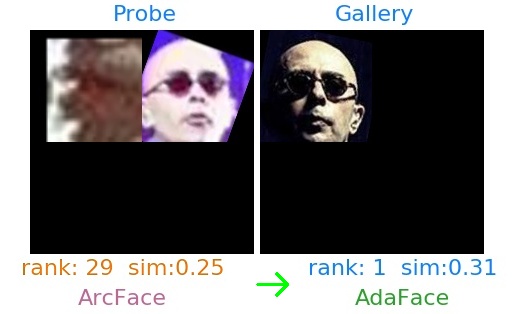}
    \end{minipage}
    \begin{minipage}{0.45\textwidth}
        \centering
        \includegraphics[width=7.6cm]{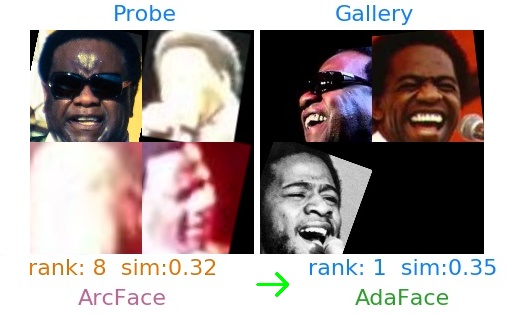}
    \end{minipage}
\end{figure}
\begin{figure}[!ht]
    \centering
    \begin{minipage}{0.45\textwidth}
        \centering
        \includegraphics[width=7.6cm]{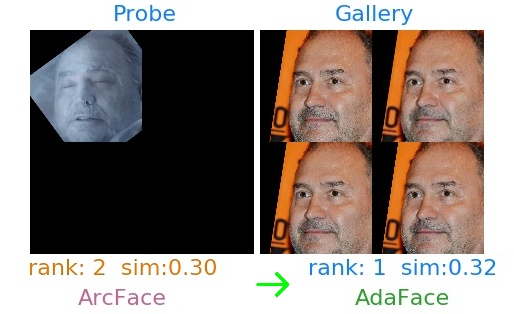}
    \end{minipage}
    \begin{minipage}{0.45\textwidth}
        \centering
        \includegraphics[width=7.6cm]{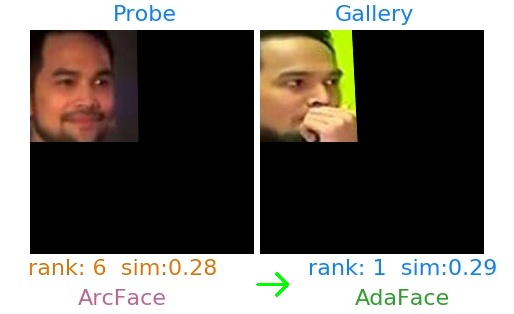}
    \end{minipage}
\end{figure}
\begin{figure}[!ht]
    \centering
    \begin{minipage}{0.45\textwidth}
        \centering
        \includegraphics[width=7.6cm]{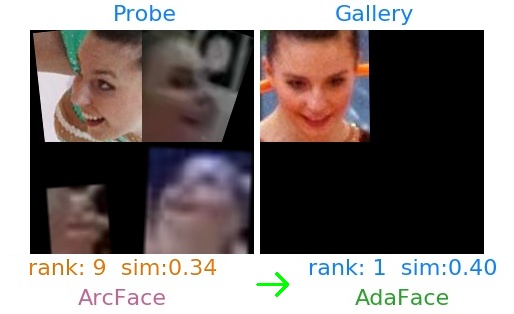}
    \end{minipage}
    \begin{minipage}{0.45\textwidth}
        \centering
        \includegraphics[width=7.6cm]{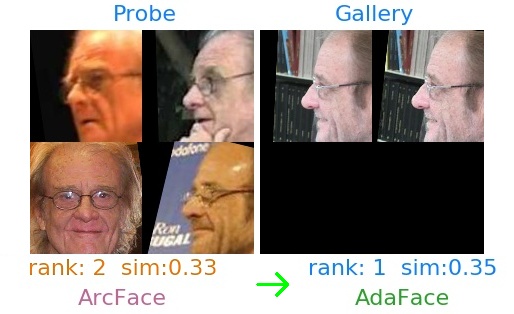}
    \end{minipage}
\end{figure}
\begin{figure}[!ht]
    \centering
    \begin{minipage}{0.45\textwidth}
        \centering
        \includegraphics[width=7.6cm]{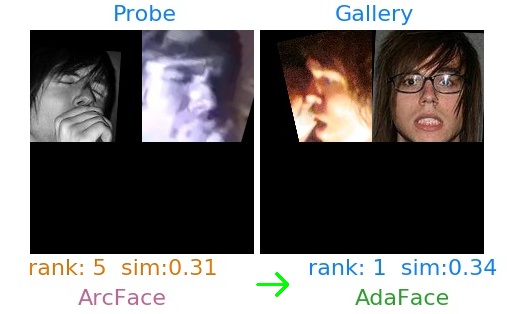}
    \end{minipage}
    \begin{minipage}{0.45\textwidth}
        \centering
        \includegraphics[width=7.6cm]{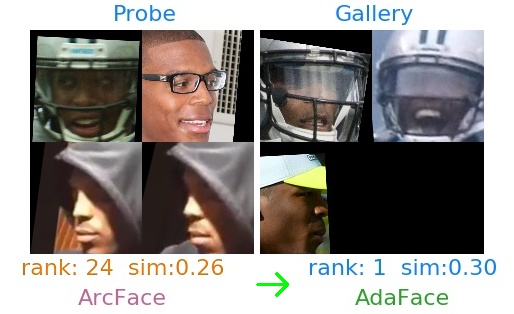}
    \end{minipage}
\end{figure}
\begin{figure}[!ht]
    \centering
    \begin{minipage}{0.45\textwidth}
        \centering
        \includegraphics[width=7.6cm]{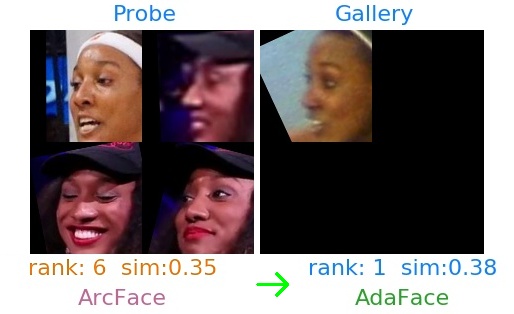}
    \end{minipage}
    \begin{minipage}{0.45\textwidth}
        \centering
        \includegraphics[width=7.6cm]{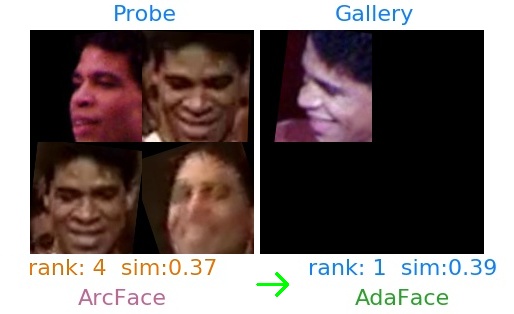}
    \end{minipage}
\end{figure}
\begin{figure}[!ht]
    \centering
    \begin{minipage}{0.45\textwidth}
        \centering
        \includegraphics[width=7.6cm]{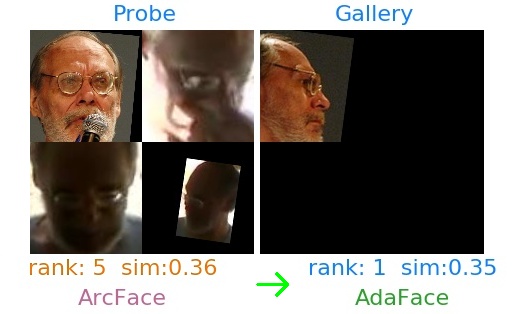}
    \end{minipage}
    \begin{minipage}{0.45\textwidth}
        \centering
        \includegraphics[width=7.6cm]{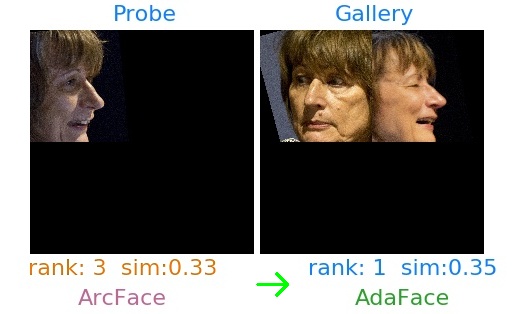}
    \end{minipage}
    \caption{Examples from IJB-C~\cite{ijbc} dataset, where ArcFace fails to identify the subject whereas AdaFace successfully finds the correct match between the probe and the gallery. On the left is the set of probe images and on the right is the set of gallery images. }
\end{figure}

\clearpage
\section{Comparison with General Image-Quality Aware Learning Method}

We compare our method with QualNet~\cite{kim2021quality} (CVPR21) as a comparison with general image-quality aware learning method. The scope of general image-quality aware learning methods is not limited to face recognition, but the idea is applicable. 
In Tab.~\ref{rebut:table1}, we show the comparison with QualNet with models trained on CASIA-WebFace. AdaFace outperforms QualNet on the TinyFace test set.
QualNet aligns the low quality (LQ) image feature distribution to the high quality (HQ) features' distribution via a fixed pretrained decoder.
In contrast, AdaFace prevents LQ images from degrading the overall recognition performance by de-emphasizing heavily degraded LQ images. Since LQ facial images can often be devoid of identity, it helps to avoid overfitting on unidentifiable LQ images and learn to exploit the identifiable LQ images. This improves generalization across HQ and LQ.

\begin{table}[h]
\centering
\small
\setlength{\tabcolsep}{3pt}
\begin{tabular}{|c|c|c||c|c|}
\hline
Method & Training Set        & Test set            & Rank1    & Rank5    \\ \hline\hline
QualNet~\cite{kim2021quality} &   \multirow{2}{*}{CASIA-Webface}   & \multirow{2}{*}{TinyFace}  &  $35.54$ &   $44.45$         \\
AdaFace &     &   &  \bm{$44.39$}  &   \bm{$47.23$}         \\ \hline
\end{tabular}
\caption{Closed set identification performance (ranked match rate) on TinyFace. For a fair comparison, we adopt the train/test setting of QualNet. QualNet results are directly taken from the CVPR21 paper.}
\label{rebut:table1}
\end{table}

\section{Effect of Batch Size}
Our image quality proxy $\widehat{\Vert \bm{z}_i \Vert}$ does not depend on the batch size due to the exponential moving average in Eq.17 of the main paper (rewritten below). 
\begin{equation}
    \widehat{\Vert \bm{z}_i \Vert} = \clip{\frac{\Vert \bm{z}_i \Vert - \mu_{z}}{\sigma_{z} / h}}^1_{-1},
\end{equation}
\begin{equation}
    \mu_z = \alpha \mu_z^{(k)} + (1-\alpha) \mu_z^{(k-1)}.
\end{equation} 
To empirically show this, we train R50 model on MS1MV2 with the batch size of $128$, $256$ and $512$ and report their performance on IJB-B TAR@FAR=0.01\%. As shown in Tab.~\ref{rebut:table2}, the difference due to the batch size is minimal. 

\begin{table}[h]
\centering
\small
\setlength{\tabcolsep}{3pt}
\begin{tabular}{|c|c|c||c|}
\hline
Method & Batch size $128$        & Batch size $256$            & Batch size $512$      \\ \hline\hline
AdaFace &   $94.32$  & $94.42$ & $94.35$
  \\ \hline
\end{tabular}
\caption{Performance comparison by varying the batch size. This shows that AdaFace performance not subject to different batch sizes.}
\label{rebut:table2}
\end{table}

\section{Implementation Details and Code}
The code is released at \url{https://github.com/mk-minchul/AdaFace}. 
For preprocessing the training data MS1MV2~\cite{deng2019arcface}, we reference InsightFace~\cite{insightface} and InsightFacePytorch~\cite{InsightFace_Pytorch}, for the backbone model definition, TFace~\cite{TFace} and for evaluation of LFW~\cite{lfw}, CFP-FP~\cite{cfpfp}, CPLFW~\cite{cplfw}, AgeDB~\cite{agedb}, CALFW~\cite{calfw}, IJB-B~\cite{ijbb}, and IJB-C~\cite{ijbc}, we use InsightFace ~\cite{insightface}. For preprocessing IJB-S~\cite{ijbs} and TinyFace~\cite{tinyface}, we use MTCNN~\cite{zhang2016joint} to align faces.

\end{document}